\documentclass[final]{nesy2026}

\let\cite\citep

\usepackage{xcolor}
\usepackage{listings}
\lstdefinestyle{asp}{
  basicstyle=\ttfamily\footnotesize,
  breaklines=true,
  columns=fullflexible,
  frame=single,
  framesep=5pt,
  commentstyle=\itshape\color{gray},
  morecomment=[l]{\%},
  extendedchars=true,
  literate=%
    {→}{$\rightarrow$}{1}
    {≡}{$\equiv$}{1}
    {—}{---}{1}
    {–}{--}{1}
    {≤}{$\leq$}{1}
    {≥}{$\geq$}{1}
    {≠}{$\neq$}{1}
    {…}{\ldots}{1}
    {é}{\'{e}}{1}
    {á}{\'{a}}{1}
    {í}{\'{\i}}{1}
    {ó}{\'{o}}{1}
    {ú}{\'{u}}{1}
    {ñ}{\~{n}}{1}
    {═}{=}{1}
    {│}{|}{1}
    {─}{-}{1}
    {├}{+}{1}
    {┤}{+}{1}
    {┬}{+}{1}
    {┴}{+}{1}
    {┼}{+}{1}
    {╔}{+}{1}
    {╗}{+}{1}
    {╚}{+}{1}
    {╝}{+}{1}
    {║}{|}{1}
    {•}{*}{1}
    {◦}{o}{1}
    {▶}{>}{1}
    {◀}{<}{1}
    {●}{*}{1}
    {○}{o}{1}
    {√}{V}{1}
    {×}{x}{1},
}

\usepackage{float}
\usepackage{booktabs}
\usepackage{multirow}
\usepackage{array}
\usepackage[skip=4pt]{caption}
\usepackage{tikz}
\usetikzlibrary{positioning, arrows.meta, fit, calc}
\usepackage{enumitem}
\setlist[enumerate]{itemsep=1pt, topsep=3pt}

\title[Distilling ASP Theories]{Distilling Answer Set Programming Theories\\ from Large Language Models}

\clearauthor{\Name{Nelson {Higuera Ruiz}} \Email{nelson.higueraruiz@extensity.ai}\\
  \Name{Markus Hofmarcher} \Email{markus@extensity.ai}\\
  \Name{Claudiu {Leoveanu-Condrei}} \Email{leo@extensity.ai}\\
  \addr ExtensityAI, Vienna, Austria}

\begin{document}

\maketitle

\begin{abstract}
Writing Answer Set Programming (ASP) theories from scratch is a difficult and time-consuming task. We take a neurosymbolic approach to study whether a large language model can distill complete and correct theories, given a fixed agent harness with the solver in the loop. The protocol is dataset-agnostic: with a single fixed prompt and an empty file as the starting point the model is given a 1-hour time limit to derive a complete theory.
We chose VQA as the application domain, across three benchmarks (CLEVR, GQA, CLEVRER), as these are publicly available and non-trivial.
In order to study the model scale required for solving this task we evaluate nine different models: four frontier (Claude Sonnet 4.6, Claude Opus 4.7, GPT-5, DeepSeek V4 Pro), two mid-tier (DeepSeek V4 Flash, gpt-oss-120b), and three smaller open-weights (qwen3.6-27b, gpt-oss-20b, qwen3.5-9b). Three of four frontier models reach $100\%$ on CLEVR and $92.8\%$--$98.8\%$ on GQA; on CLEVRER, Sonnet, Opus, and DeepSeek V4 Pro score $92.7\%$--$95.3\%$. GPT-5 reaches $98.7\%$ on CLEVR but drops to $41.8\%$ on GQA and to $86.7\%$ on CLEVRER. Adding handwritten reference theories from other datasets moves the other three frontier models by at most $\pm 3.4$\,pp but reduces GPT-5's accuracy by 3--19\,pp. We release the code, prompts, and theories distilled\footnote{Code available at \url{https://github.com/pudumagico/distilling-theories}}.
\end{abstract}

\begin{keywords}
  Answer Set Programming, Large Language Models, Neurosymbolic Reasoning, LLM Agents, Visual Question Answering
\end{keywords}

\section{Introduction}

Neurosymbolic computation~\cite{garcez2023neurosymbolic} pairs neural perception and parsing with a symbolic reasoning module. A logic programming language such as Answer Set Programming (ASP)~\cite{gelfond1988stable}, executed by a solver like \texttt{clingo}~\cite{gebser2019multishot}, gives the reasoning module attractive properties: it is declarative, inspectable, and debuggable. The cost is authoring. A theory for a non-trivial reasoning task may contain hundreds of rules covering every operator the task can invoke, and handwritten theories do not transfer across domains whose schemas differ.

We study whether a large language model can distill such a theory from scratch, given an empty file, a fixed prompt, and an agent harness with shell access to the solver. The model reads a few training examples, runs the solver on a candidate theory, identifies failure modes, edits the theory, and iterates until validation accuracy plateaus or a 1-hour time limit is reached. The unit of work is the entire theory, authored across many edit cycles. No human-supplied template, partial theory, or rule pattern is provided; the only optional input is a small number of read-only handwritten theories from \emph{other} domains.

We instantiate the protocol on Visual Question Answering (VQA)~\cite{antol2015vqa,goyal2017vqav2} as the application domain. VQA decomposes naturally into scene parsing, question parsing, and reasoning; three benchmarks span our evaluation: CLEVR~\cite{johnson2017clevr} stresses compositional reasoning over synthetic 3D scenes, GQA~\cite{hudson2019gqa} moves to real images with a substantially larger operator, attribute, and class vocabulary parsed from Visual Genome scene graphs, and CLEVRER~\cite{yi2020clevrer} introduces temporal, causal, and counterfactual reasoning over short videos. The datasets' own scene annotations and functional programs supply the perception and parsing inputs; the model authors only the reasoning theory.

The object of study is the ASP theory the model produces. We measure its quality by accuracy on a held out validation slice, with handwritten theories as reference data.

Our contributions are:

\begin{enumerate}
  \item A distillation protocol that is identical across datasets and models. The agent harness, system prompt, and tool set are fixed across configurations, only the model identifier and the dataset folder vary; each experiment is run in a sandbox. 
  \item Experimental results showing that agents can distill complete ASP theories: three of four frontier models reach the CLEVR ceiling, exceed the handwritten GQA ceiling, and score $92.7\%$--$95.3\%$ on CLEVRER. 
  \item A taxonomy of observed failure modes
  and an analysis of how reference theories interact with model capability and dataset difficulty.
\end{enumerate}

\section{Related Work}

\paragraph{LLMs writing logic programs.}
\citet{yang2023coupling,ishay2023llmasp,yang2024learning} translate natural-language sentences into answer-set programs for textual question answering; \citet{trinh2024alphageometry} synthesize olympiad-geometry proofs in a symbolic system from LLM-generated programs. Closer to our setting, \citet{ren2025aspbenchmark} benchmark LLMs on writing ASP programs and report substantial model-by-model variation in ASP authoring competence, consistent with what we observe in the smaller-model controls. \citet{coppolillo2026asp} fine-tune LLMs for ASP; we leave the model unchanged and rely on the agent harness. \citet{schrader2026solverinloop} couple an LLM to a solver in a feedback loop for ASP logic-puzzle solving; their unit of work is a single-shot answer, ours is a full theory authored across many edit cycles. \citet{borroto2025qa} pair an LLM with inductive learning of ASP rules from answer sets for commonsense story QA; theirs is rule-by-rule induction, ours is whole-theory authoring under a fixed schema.
The closest prior work in our application is \citet{eiter2024declarative}, which distills logic rules from an LLM one rule at a time, prompting the model separately for each question shape and assembling the program from the pieces. We differ in two ways. First, the unit of work is the entire theory, authored across many edit cycles with the solver in the loop, not a single rule. Second, the model is told nothing about the target theory's structure, so there is no per-question-shape template to populate. The classical knowledge-distillation framework~\cite{hinton2015distilling} transfers continuous teacher outputs to a student model; we transfer categorical answers via an intermediate symbolic theory.

\paragraph{Neurosymbolic VQA.}
NS-VQA~\cite{yi2018neural} established the three-stage perception/parser/inference template that organizes most subsequent work. NS-CL~\cite{mao2019neuro} closes the loop by learning visual concepts, word meanings, and the semantic parser jointly from question--answer pairs, with no explicit supervision on concept labels or program traces. Neural module networks~\cite{andreas2016neural}, program execution~\cite{johnson2017inferring}, and compositional attention networks~\cite{hudson2018mac} are nearby points in the design space; the Garcez and Lamb survey~\cite{garcez2023neurosymbolic} situates them in the wider neurosymbolic landscape. DeepProbLog~\cite{manhaeve2018deepproblog} and Logic Tensor Networks~\cite{donadello2017ltnvqa} integrate neural perception with probabilistic logic but assume a fixed inference program; our work targets the program itself.

\paragraph{LLM agents.}
Agent harnesses operate the LLM through a shell with file and command tools: SWE-bench~\cite{jimenez2024swebench} and SWE-agent~\cite{yang2024sweagent} on real-world bug-fix tasks, AutoGen~\cite{wu2023autogen} for multi-agent composition, Voyager~\cite{wang2023voyager} for open-ended exploration. We use OpenCode\footnote{\url{https://opencode.ai}}, which exposes the same surface, and apply it to theory authoring. Inference-time reasoning techniques (chain-of-thought~\cite{wei2022cot,kojima2022zeroshot}, self-consistency~\cite{wang2023selfconsistency}, tree-of-thoughts~\cite{yao2023tot}, program-aided language models~\cite{gao2023pal,chen2023pal}) elicit step by step reasoning from a frozen LLM; our setting differs in that the deliverable is the ASP theory itself, not the chain of thought.

\section{Background}

\subsection{Answer Set Programming}

An ASP program is a set of rules $h \leftarrow b_1, \dots, b_k, \mathrm{not}\ b_{k+1}, \dots, \mathrm{not}\ b_n$. A stable model (answer set) is a minimal model under the semantics of \citet{gelfond1988stable}; we use \texttt{clingo}~\cite{gebser2019multishot} to enumerate them. Each question is answered by deriving \texttt{ans/1} atoms (the ASP notation \texttt{p/n} marks predicate \texttt{p} of arity \texttt{n}; \texttt{ans/1} is unary, with the answer as its single argument). Open-ended answers (colors, names, counts) are scored under brave entailment; yes/no answers under a stricter rule that requires the derived set to be a singleton with the correct value.

A standard neurosymbolic VQA pipeline has three stages: a perception module that emits scene facts, a semantic parser that compiles the question into a deterministic program, and an inference engine that executes the program against the scene to obtain the answer~\cite{yi2018neural,mao2019neuro}. Our protocol replaces the third stage: one \texttt{clingo} invocation receives scene facts $s_\mathrm{asp}$, question facts $q_\mathrm{asp}$ for a question over that scene, and the theory $T$, and enumerates the stable models of $s_\mathrm{asp} \cup q_\mathrm{asp} \cup T$. The model's task is to author rules that interpret each operator emitted by the parser.

\subsection{Datasets}

\begin{figure}[t]
\centering
\begin{minipage}[t]{0.30\textwidth}\centering
  \includegraphics[width=\linewidth,height=2.6cm,keepaspectratio]{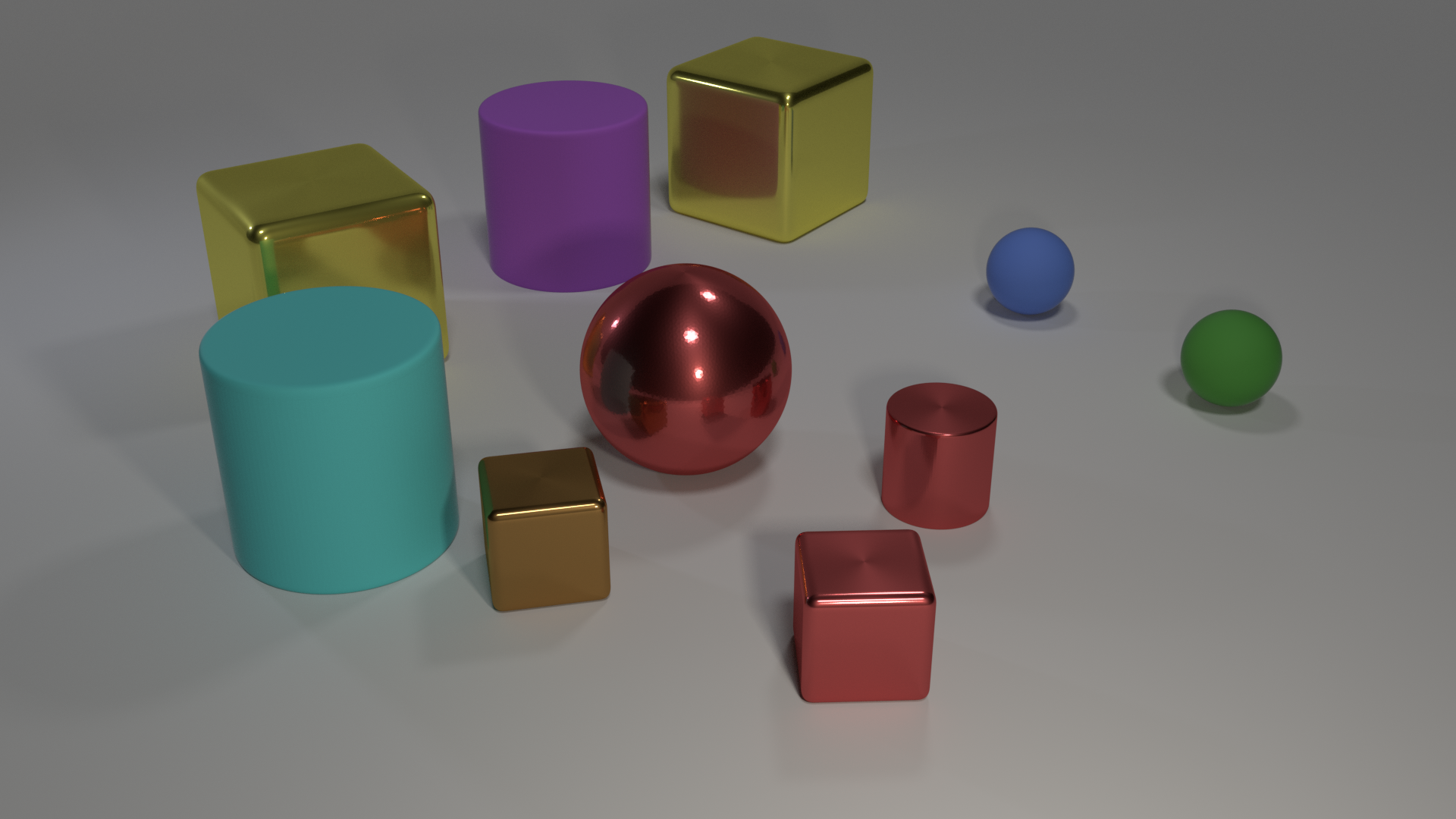}\\
  \small (a) CLEVR
\end{minipage}\hfill
\begin{minipage}[t]{0.30\textwidth}\centering
  \includegraphics[width=\linewidth,height=2.6cm,keepaspectratio]{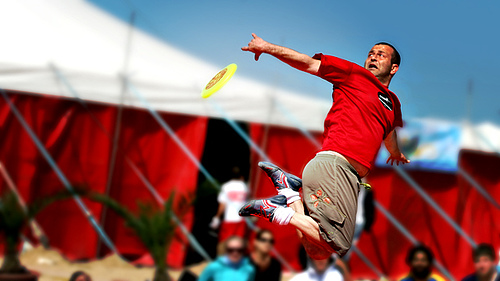}\\
  \small (b) GQA
\end{minipage}\hfill
\begin{minipage}[t]{0.30\textwidth}\centering
  \includegraphics[width=\linewidth,height=2.6cm,keepaspectratio]{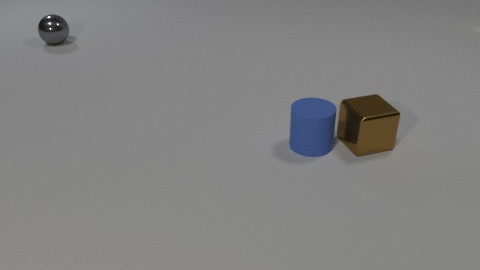}\\
  \small (c) CLEVRER (video frame)
\end{minipage}
\caption{Representative examples of the three datasets.}
\label{fig:datasets}
\end{figure}

The three benchmarks we target are shown in Figure~\ref{fig:datasets}. \textbf{CLEVR}~\cite{johnson2017clevr} renders synthetic 3D scenes of up to ten geometric primitives with compositional questions (count, exist, attribute query, spatial relate, set operations); the handwritten reference theory has $64$ rules and reaches $100\%$ on val/200. \textbf{GQA}~\cite{hudson2019gqa} uses real images parsed from Visual Genome~\cite{krishna2017visualgenome} scene graphs, with questions over a substantially larger vocabulary of object classes, attributes, and binary relations; the reference theory has $57$ rules and reaches $77.5\%$. \textbf{CLEVRER}~\cite{yi2020clevrer} uses short videos of colliding rigid bodies with descriptive, predictive, counterfactual, and explanatory question categories; to the best of our knowledge, no handwritten ASP theory is publicly available. Each dataset ships ground truth scene and question annotations that can be transformed into ASP facts $s_\mathrm{asp}$ and $q_\mathrm{asp}$.

Figure~\ref{fig:worked-example} walks through one GQA example, taking the scene in Figure~\ref{fig:datasets}b (middle) as the source. The parser emits the scene facts $s_\mathrm{asp}$ and question facts $q_\mathrm{asp}$; the handwritten reference theory provides the rules that interpret each operator. Concatenating the three and running \texttt{clingo} yields \texttt{ans(yes)} as the unique answer atom.

\begin{figure}[ht]
\small
\noindent\rule{\linewidth}{0.4pt}
\vspace{2pt}
\noindent\begin{minipage}[t]{0.48\linewidth}
\textbf{Scene + question facts (parser output):}
\begin{lstlisting}[style=asp,frame=none]
% scene: man + yellow frisbee
object(2718).
has_attr(2718, class, frisbee).
has_attr(2718, color, yellow).
object(3141).
has_attr(3141, class, person).
% question: "Is the frisbee yellow?"
scene(0).
select(1, 0, frisbee).
verify_attr(2, 1, color, yellow).
end(2).
\end{lstlisting}
\end{minipage}\hfill\begin{minipage}[t]{0.48\linewidth}
\textbf{Excerpt from the handwritten theory $T$:}
\begin{lstlisting}[style=asp,frame=none]
state(TO, ID) :- scene(TO), object(ID).
state(TO, ID) :- select(TO, TI, CLASS),
  state(TI, ID), has_attr(ID, class, CLASS).
bool(TO, yes) :- verify_attr(TO, TI, ATTR, VALUE),
  state(TI, ID), has_attr(ID, ATTR, VALUE).
bool(TO, no)  :- verify_attr(TO, TI, ATTR, VALUE),
  not bool(TO, yes).
ans(V) :- end(TO), bool(TO, V).
\end{lstlisting}
\end{minipage}
\vspace{2pt}
\noindent\rule{\linewidth}{0.4pt}
\caption{One GQA example end-to-end, against the scene in Figure~\ref{fig:datasets}b. The solver concatenates scene and question facts with the theory $T$ and computes stable models; \texttt{clingo} derives \texttt{ans(yes)} as the unique answer atom. The theory excerpt is verbatim from the handwritten GQA reference; the LLM's task is to author rules of this shape for every operator the parser emits.}
\label{fig:worked-example}
\end{figure}

\section{Distillation Protocol}

\begin{figure}[t]
\centering
\includegraphics[width=0.92\linewidth]{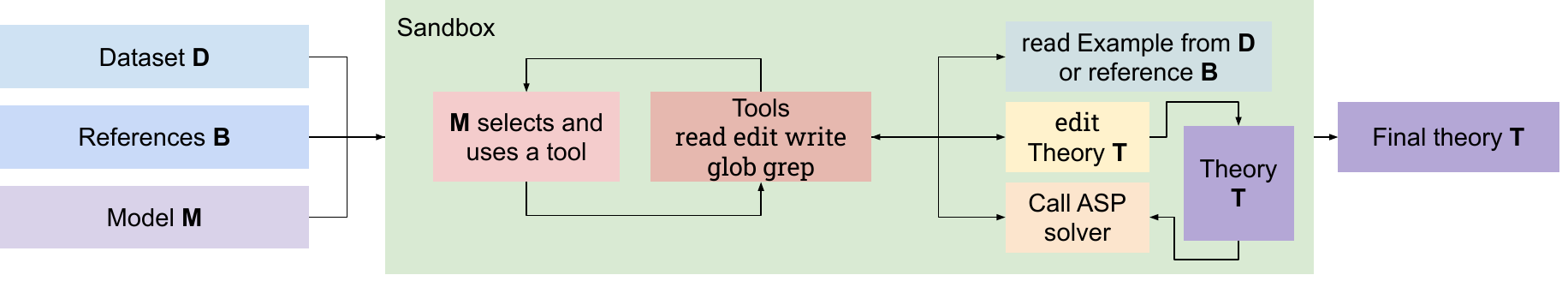}
\caption{System overview. Three inputs (dataset $D$, $B$ reference theories from other datasets, model $M$) enter a sandbox in which $M$ iteratively picks a tool (\texttt{read}, \texttt{edit}, \texttt{write}, \texttt{glob}, \texttt{grep}) to read a training example or reference, edit the theory~$T$, or call the ASP solver on $T$. The loop runs until $M$ self-stops or the 1-hour cap fires; the output is the final theory~$T$, scored on a held-out validation set outside the sandbox.}
  \label{fig:system}
\end{figure}

Figure~\ref{fig:system} gives an overview of the system. A configuration fixes a dataset, a reference count $B$, and a model (CLEVRER $B = 1$ further fixes which reference is shown). A \emph{sample} is one run of the agent under a configuration, producing one final \texttt{theory}. The rest of this section describes the agent's tools, the harness (including the per-sample Docker sandbox), and the per-sample algorithm; experimental specifics ($N$, 1-hour cap, model and dataset choices) are in Section~\ref{sec:setup}.

\subsection{What the model can do}
\label{sec:tools}
The model has file tools (\texttt{read}, \texttt{edit}, \texttt{write}, \texttt{glob}, \texttt{grep}) and a \texttt{bash} tool restricted to \texttt{uv run solve} and \texttt{uv run lint}. \texttt{uv run solve --idx <i> --split train} runs \texttt{clingo} on the current \texttt{theory} plus training example \texttt{i}'s facts and returns the derived \texttt{ans/1} atoms. \texttt{uv run lint} parse-checks the theory. Evaluation examples are staged outside the sandbox and never visible to the model; the final theory is scored on the validation set after the loop, from outside the container.

\subsection{Harness}
\label{sec:harness}
We use OpenCode\footnote{\url{https://opencode.ai}.}, an open source agent CLI, invoked once per sample in non-interactive mode. Each sample runs in its own Docker container (image \texttt{aspdist-sandbox}) that bind-mounts the sample's run folder as the only writable filesystem, plus the project source read only; the model's tools (Section~\ref{sec:tools}) execute inside this container. The model has no visibility into other samples, other configurations, or the dataset's own handwritten theory. OpenCode routes the conversation between the container and the LLM provider over the tool use API and exits when the model stops or the time limit is reached. The harness is identical across all samples; only the model identifier and the dataset folder change. All runs are autonomous: there is no human input during the session.

\subsection{Algorithm}

Per sample, the harness sends the prompts (Listings~\ref{lst:system}, \ref{lst:directive}; full text in Appendix~\ref{app:prompt}) and runs Algorithm~\ref{alg:loop} until the model self-declares done or the 1-hour time limit is reached. We report mean and standard deviation across the $N$ samples per configuration.

\noindent\begin{minipage}[t]{0.46\linewidth}
\begin{lstlisting}[caption={System prompt (excerpt).},label=lst:system,basicstyle=\ttfamily\scriptsize,frame=single,framesep=6pt,xleftmargin=0pt,xrightmargin=0pt,aboveskip=0pt,belowskip=2pt]
You are an ASP agent. Produce theory.lp
such that clingo derives ans(A) for each
training example. Your theory is scored on
a held out val set you cannot see; iterate
against the training examples.

Tools: read, edit, write, glob, grep, plus
bash restricted to:
uv run solve, uv run lint.
\end{lstlisting}

\vspace{2mm}

\begin{lstlisting}[caption={Initial directive (verbatim).},label=lst:directive,basicstyle=\ttfamily\scriptsize,frame=single,framesep=6pt,xleftmargin=0pt,xrightmargin=0pt,aboveskip=2pt,belowskip=2pt]
Begin. Work efficiently. Inspect only a few
training examples, write a syntactically
valid initial theory quickly, test it with
uv run solve --run-dir . --idx 0 --split train,
and iterate. Do not spend many turns reading
examples before writing the first theory.
\end{lstlisting}
\end{minipage}\hfill\begin{minipage}[t]{0.50\linewidth}
\begin{algorithm2e}[H]
\footnotesize
\SetAlgoLined
\DontPrintSemicolon
\KwIn{dataset $D$, model $M$, reference count $B \in \{0,1,2\}$, system prompt $P_s$, directive $P_d$, time budget $\tau$, sample count $N$}
\KwOut{accuracies $(a_1, \dots, a_N)$}
\For{$i \gets 1$ \KwTo $N$}{
  materialize sample $i$ on disk: empty \texttt{theory.lp}$_i$, train/val from $D$ at fresh seed $\sigma_i$, $B$ ref theories\;
  history $\gets (P_s, P_d)$\;
  \While{\textnormal{elapsed time} $\leq \tau$}{
    $u \gets M.\textsc{step}(\text{history})$ \tcp*{one tool call}
    execute $u$; append result to history\;
    \If{$M$ stopped}{\textbf{break}\;}
  }
  $a_i \gets \textsc{eval\_val}(\texttt{theory.lp}_i)$\;
}
\Return $(a_1, \dots, a_N)$, \texttt{theory.lp}
\caption{Per-configuration loop. We use $N = 3$ for all experiments.}
\label{alg:loop}
\end{algorithm2e}
\end{minipage}

The prompt is intentionally minimal: it names the task (raise \texttt{acc\_strict} on val), the tool surface (one tool call per turn; \texttt{Edit} and \texttt{Write} primary), and the pacing rule (write a valid theory quickly, evaluate as soon as it parses). It includes no ASP primer; the model is assumed to know clingo syntax. The same two prompts drive every configuration, so observed differences reflect model capability, not prompt engineering.

\section{Experiments}
\label{sec:experiments}

The experiments address four research questions.

\begin{enumerate}
  \item[\textbf{RQ1.}] Can an agent author a complete ASP theory that approaches or matches the validation accuracy of a handwritten reference theory? (Section~\ref{sec:baseline})
  \item[\textbf{RQ2.}] How does access to handwritten reference theories from other datasets ($B \in \{0, 1, 2\}$) affect the distilled theory's accuracy? (Section~\ref{sec:baseline})
  \item[\textbf{RQ3.}] At what model scale does the protocol become functional, and what failure modes appear below that scale? (Section~\ref{sec:smaller-models})
  \item[\textbf{RQ4.}] How do models differ in tool use behavior and per-example failure modes? (Section~\ref{sec:baseline}, Appendix~\ref{app:failures})
\end{enumerate}

\subsection{Setup}
\label{sec:setup}
For each configuration and model we run $N = 3$ samples with independent, time-derived seeds; each seed draws a fresh 100-train + 200-val split from a pool $10\times$ that size, so the three samples in a configuration use three different splits. The train examples are placed inside the sample's Docker sandbox (Section~\ref{sec:harness}); val examples are staged outside and used only for scoring after the loop. Each sample is capped at 1 hour; all experiments use OpenCode as the agent harness. We evaluate nine models grouped by capability tier: four frontier (Claude Sonnet 4.6, Claude Opus 4.7, GPT-5, DeepSeek V4 Pro), two mid-tier (DeepSeek V4 Flash, \texttt{gpt-oss-120b}), and three smaller open-weights (\texttt{qwen3.6-27b}, \texttt{gpt-oss-20b}, \texttt{qwen3.5-9b}); all models other than Anthropic and OpenAI are accessed via OpenRouter (Appendix~\ref{app:models}). Only CLEVR and GQA have handwritten theories, so $B \in \{0, 1\}$ for them and $B \in \{0, 1, 2\}$ for CLEVRER; the dataset's own theory is never given as a reference.

\subsection{Results}
\label{sec:baseline}

Table~\ref{tab:results} reports the 9 models $\times$ 8 configurations. The baseline ($B = 0$) is the strictest setting: no references, no ASP primer, empty \texttt{theory}. $B = 1$ adds one read only handwritten reference from another dataset; $B = 2$ adds both (CLEVRER only). On CLEVRER, $B = 1$ further splits by which reference is shown.

\begin{table}[ht]
\centering
\footnotesize
\setlength{\tabcolsep}{1pt}
\begin{tabular}{@{}l cc cc cccc@{}}
\toprule
 & \multicolumn{2}{c}{CLEVR} & \multicolumn{2}{c}{GQA} & \multicolumn{4}{c}{CLEVRER} \\
\cmidrule(lr){2-3} \cmidrule(lr){4-5} \cmidrule(lr){6-9}
Model & $B_0$ & $B_1$ & $B_0$ & $B_1$ & $B_0$ & $B_{1c}$ & $B_{1g}$ & $B_2$ \\
\midrule
Sonnet 4.6      & \textbf{100}\,$\pm$\,0    & \textbf{100}\,$\pm$\,0    & \textbf{98.8}\,$\pm$\,0.8 & \textbf{98.5}\,$\pm$\,0.4 & 95.2\,$\pm$\,2.1          & \textbf{96.0}\,$\pm$\,0.5 & 94.8\,$\pm$\,1.6          & 95.3\,$\pm$\,1.5 \\
Opus 4.7        & \textbf{100}\,$\pm$\,0    & \textbf{100}\,$\pm$\,0    & 97.5\,$\pm$\,1.2          & 96.2\,$\pm$\,2.1          & \textbf{95.3}\,$\pm$\,0.2 & 95.5\,$\pm$\,0.8          & \textbf{96.7}\,$\pm$\,1.2 & \textbf{95.5}\,$\pm$\,1.2 \\
GPT-5           & 98.7\,$\pm$\,1.9          & 88.8\,$\pm$\,8.5          & 41.8\,$\pm$\,9.8          & 35.2\,$\pm$\,6.3          & 86.7\,$\pm$\,4.7          & 77.3\,$\pm$\,10.2         & 83.5\,$\pm$\,7.9          & 67.3\,$\pm$\,13.2 \\
DeepSeek V4 Pro & \textbf{100}\,$\pm$\,0    & \textbf{100}\,$\pm$\,0    & 92.8\,$\pm$\,1.2          & 96.2\,$\pm$\,2.2          & 92.7\,$\pm$\,0.6          & 95.2\,$\pm$\,2.2          & 94.7\,$\pm$\,1.6          & 95.2\,$\pm$\,0.9 \\
\midrule
DeepSeek V4 Flash & \textbf{100}\,$\pm$\,0  & 98.0\,$\pm$\,2.0          & 85.0\,$\pm$\,1.5          & 27.5\,$\pm$\,47.6         & 58.3\,$\pm$\,50.6         & 86.3\,$\pm$\,5.5          & 11.0\,$\pm$\,19.1         &  0.0\,$\pm$\,0 \\
gpt-oss-120b    &  0.0\,$\pm$\,0            &  0.0\,$\pm$\,0            &  0.0\,$\pm$\,0            &  0.0\,$\pm$\,0            &  0.0\,$\pm$\,0            &  0.0\,$\pm$\,0            &  0.0\,$\pm$\,0            &  0.0\,$\pm$\,0 \\
qwen3.6-27b     & 42.2\,$\pm$\,49.8         & 66.7\,$\pm$\,57.7         & 85.0\,$\pm$\,6.2          & 77.7\,$\pm$\,9.7          & 39.8\,$\pm$\,34.5         &  0.0\,$\pm$\,0            & 12.0\,$\pm$\,20.8         &  0.0\,$\pm$\,0 \\
gpt-oss-20b     &  0.0\,$\pm$\,0            &  0.0\,$\pm$\,0            &  0.0\,$\pm$\,0            &  0.0\,$\pm$\,0            &  0.0\,$\pm$\,0            &  0.0\,$\pm$\,0            &  0.0\,$\pm$\,0            & 11.8\,$\pm$\,20.5 \\
qwen3.5-9b      & 17.7\,$\pm$\,25.0         &  0.0\,$\pm$\,0            &  0.0\,$\pm$\,0            &  0.0\,$\pm$\,0            &  0.5\,$\pm$\,0.7          &  0.0\,$\pm$\,0            &  0.0\,$\pm$\,0            &  0.0\,$\pm$\,0 \\
\bottomrule
\end{tabular}
\caption{\texttt{acc\_strict} on val/200 (\%), mean $\pm$ std across $n = 3$ samples per configuration. $B \in \{0, 1, 2\}$ is the reference-theory count; for
CLEVRER $B = 1$, the choice between CLEVR-ref ($B_{1c}$) and GQA-ref ($B_{1g}$) is itself an experimental condition. Frontier models above the midrule, sub-frontier below.}
\label{tab:results}
\end{table}

\paragraph{Baseline ($B = 0$).} Three frontier models reach $100\%$ on CLEVR and GPT-5 reaches $98.7\%$. On GQA, Sonnet, Opus, and DeepSeek V4 Pro score $92.8\%$--$98.8\%$ while GPT-5 drops to $41.8\%$ (the theory derives no \texttt{ans/1} for most val examples; see Appendix~\ref{app:failures}). On CLEVRER the same three score $92.7\%$--$95.3\%$; GPT-5 reaches $86.7\%$.

\paragraph{Reference ablation ($B \geq 1$).} Sonnet, Opus, and DeepSeek V4 Pro stay within $\pm 3.4$\,pp across $B$ (DeepSeek GQA $B = 0 \to B = 1$, $+3.4$\,pp; all other movements smaller). GPT-5 regresses on every configuration: $-9.9$ (CLEVR), $-6.6$ (GQA), $-9.4 / -3.2$ (CLEVRER $B = 1$ CLEVR-/GQA-ref), $-19.4$ ($B = 2$). On CLEVRER, GPT-5 scores $77.3\% \pm 10.2$ with the CLEVR-ref and $83.5\% \pm 7.9$ with the GQA-ref; the difference is within sample noise. The other three are within sample noise across the two refs. We do not have a confirmed mechanism for GPT-5's regression; one observation: GPT-5 writes shorter theories when given a reference (15--20 head predicates on CLEVRER vs.\ 43 without), so the reference may consume context budget that would otherwise go to authoring.

\paragraph{Sub-frontier.} DeepSeek V4 Flash reaches $100\%$ on CLEVR and $85.0\%$ on GQA $B = 0$; \texttt{qwen3.6-27b} reaches $85.0\%$ on GQA $B = 0$ and is high-variance on CLEVR (one
sample at $100\%$, two at $0\%$, mean $42.2 \pm 49.8$). \texttt{gpt-oss-20b} fails on tool call format; \texttt{gpt-oss-120b} fails differently (sessions terminate after 1--2 tool calls); \texttt{qwen3.5-9b} rarely emits valid clingo (only $7/19$ samples parse). Both Flash and qwen3.6-27b degrade on CLEVRER, especially when given a reference.

\begin{figure}[ht]
\centering
\includegraphics[width=0.92\linewidth]{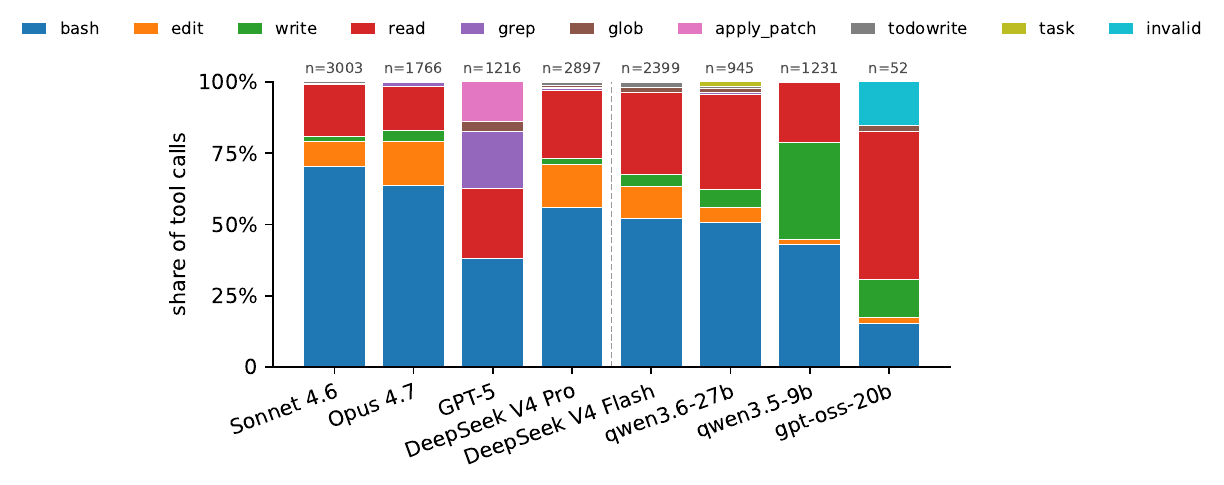}
\caption{Tool use distribution per model, summed across every completed sample. Each cluster is one tool; bar height is the fraction of that model's total tool calls. Tools with $< 1\%$ usage are dropped. \texttt{gpt-oss-120b} is omitted (every sample emits a single tool call before the session aborts). Claude pairs heavy \texttt{bash} (clingo introspection) with line-level \texttt{edit}; GPT-5 uses \texttt{apply\_patch} whole-file rewrites and \texttt{grep}-heavy navigation; the DeepSeek family matches Claude's pattern at lower volume; the smaller open-weights show varied profiles (Section~\ref{sec:smaller-models}). The same \texttt{acc\_strict} is reachable from any profile.}
\label{fig:tools}
\end{figure}

\paragraph{Tool use and failure modes.} Frontier tool-use profiles (Figure~\ref{fig:tools}) split into three patterns: Claude pairs heavy \texttt{bash} with line-level \texttt{edit}; GPT-5 uses \texttt{apply\_patch} (whole-file rewrites) and \texttt{grep}-heavy navigation; DeepSeek V4 Pro matches Claude's pattern at lower volume. Per-val-example failures fall into four classes, \textsc{parse} (the theory does not parse) and \textsc{ground} (it parses but cannot ground; folded with \textsc{parse} in Table~\ref{tab:failures} but broken out in Appendix~\ref{app:non-agent}), \textsc{no\_answer} (it parses, grounds, but derives no \texttt{ans/1}), and \textsc{semantic} (it derives a wrong answer). 
Per-model rates (Table~\ref{tab:failures}) put three frontier models at $2.5$--$4.2\%$ total error; GPT-5 at $27.6\%$; DeepSeek V4 Flash at $41.7\%$; the smaller open-weights at $57$--$95\%$; and \texttt{gpt-oss-120b} at $100\%$ (entirely \textsc{no\_answer}, since no theory ever derives an \texttt{ans/1}). Every cataloged session-level failure mode fires zero times for the four frontier models; the five sub-frontier models exhibit \textsc{runaway exploration}, \textsc{parse-broken}, \textsc{dialect mismatch}, and \textsc{early stop} (Appendix~\ref{app:failures}).

\begin{figure}[ht]
\centering
\includegraphics[width=0.92\linewidth]{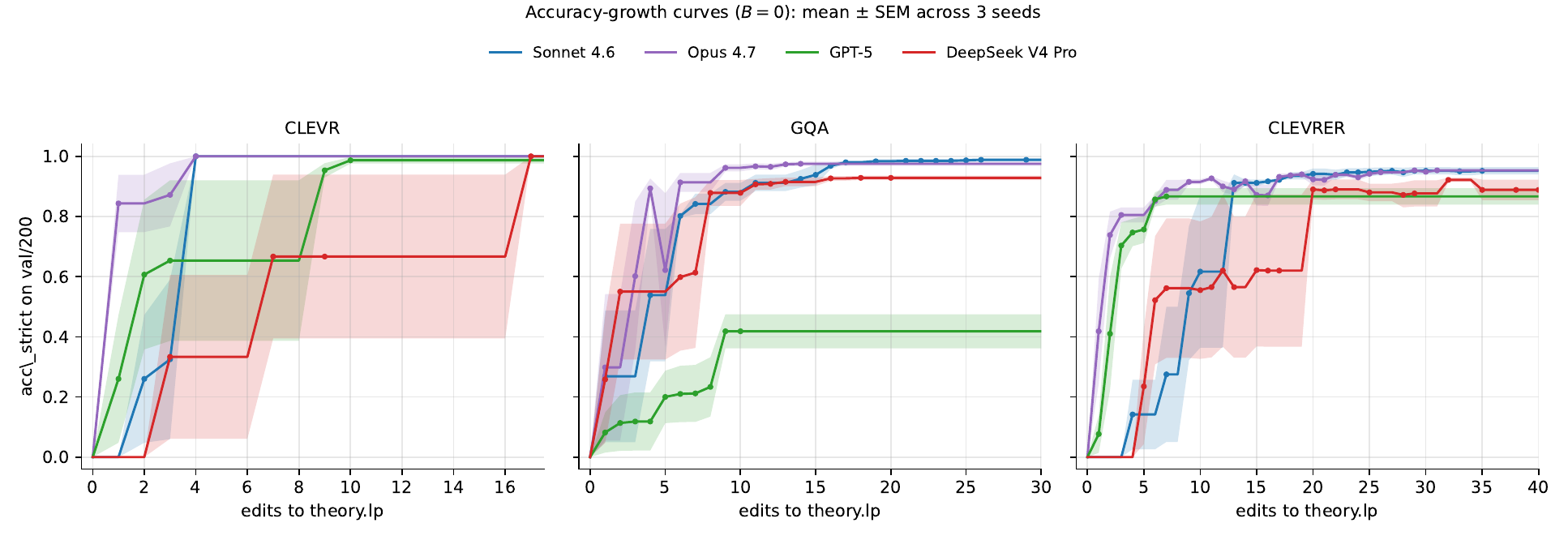}
\caption{Accuracy-growth curves: $\texttt{acc\_strict}$ on val/200 as a function of edits to \texttt{theory}, $B = 0$ condition, frontier models only.
Each curve is the mean across 3 samples; the shaded band is $\pm 1$~SEM (standard error of the mean, $= \text{sample std} / \sqrt{3}$). CLEVR (left) is solved by every model within a handful of edits. GQA (center) splits Sonnet, Opus, and DeepSeek (top band, $\geq 92\%$) from GPT-5 (lower band, $\sim 42\%$). CLEVRER (right) shows three models converging in a tight $93\%$--$95\%$ band; GPT-5 stays $\sim 87\%$.}
\label{fig:acc-growth}
\end{figure}

\begin{table}[ht]
\begin{minipage}[t]{0.55\linewidth}
\centering
\footnotesize
\setlength{\tabcolsep}{2pt}
\begin{tabular}{@{}l c c c c@{}}
\toprule
Model & \textsc{parse} & \textsc{no\_answer} & \textsc{semantic} & total \\
\midrule
Sonnet 4.6        &  0.0 &  0.2 &  2.3 &  2.5 \\
Opus 4.7          &  0.0 &  0.5 &  2.4 &  2.9 \\
GPT-5             &  0.0 & 13.5 & 14.0 & 27.6 \\
DeepSeek V4 Pro   &  0.0 &  0.8 &  3.4 &  4.2 \\
\midrule
DeepSeek V4 Flash &  4.2 & 32.2 &  5.3 & 41.7 \\
gpt-oss-120b      &  0.0 &100.0 &  0.0 &100.0 \\
qwen3.6-27b       &  0.0 & 53.1 &  3.5 & 56.6 \\
gpt-oss-20b       &  4.2 & 79.2 & 11.4 & 94.7 \\
qwen3.5-9b        & 33.3 & 43.1 & 17.3 & 93.8 \\
\bottomrule
\end{tabular}
\captionof{table}{Per-val-example failure rates.
Categories defined in Section~\ref{sec:baseline} (\textsc{parse} and \textsc{ground} folded here). \texttt{total} sums the three; the implied \texttt{acc\_strict} is $100\% - \text{total}$.}
\label{tab:failures}
\end{minipage}\hfill
\begin{minipage}[t]{0.43\linewidth}
\centering
\footnotesize
\setlength{\tabcolsep}{1.5pt}
\begin{tabular}{@{}l r r r@{}}
\toprule
Model & CLEVR & GQA & CLEVRER \\
\midrule
Sonnet 4.6       &  70 & 175 &  81 \\
Opus 4.7         &  88 & 343 & 103 \\
GPT-5            & 111 &  65 &  64 \\
DeepSeek V4 Pro  & 132 & 274 &  70 \\
\midrule
DeepSeek V4 Flash& 264 & 237 & 351 \\
gpt-oss-120b     & --- & --- & --- \\
qwen3.6-27b      &  85 & 105 &  62 \\
gpt-oss-20b      &  23 & --- &   1 \\
qwen3.5-9b &  69 &  79 &  48 \\
\midrule
\textit{handwritten} & 64 & 57 & --- \\
\bottomrule
\end{tabular}
\captionof{table}{Theory size (rules) per model at $B = 0$, mean over samples that produced a non-empty \texttt{theory}. 
}
\label{tab:theory_size}
\end{minipage}
\end{table}

\subsection{Sample distilled rules and theory size}
\label{sec:theory-size}

Theories vary substantially in length (Table~\ref{tab:theory_size}). GPT-5 writes only $65$ rules for GQA --- a fifth of what Opus writes ($343$), and close in size to the handwritten reference ($57$). The two short theories cover different question
shapes, though: GPT-5's per-val-example failures (Table~\ref{tab:failures}) are dominated by \textsc{no\_answer} rather than \textsc{semantic} errors, consistent with under-coverage.
Rule count is not coverage. At the ceiling, theory length is authoring style, not coverage: on CLEVR, Sonnet reaches $100\%$ with $70$ rules while DeepSeek Pro does the same with $132$; on GQA, Opus writes 343 rules while Sonnet writes 175, both above 97\%.
Appendix~\ref{app:rules-sample} reproduces a representative GQA fragment.

\subsection{Sub-frontier controls}
\label{sec:smaller-models}

Five sub-frontier models exhibit three failure modes (Table~\ref{tab:results}; details in Appendix~\ref{app:failures}). \textbf{(i)~Tool call format.} \texttt{gpt-oss-20b} emits malformed tool names the harness rejects on $17/24$ samples. \texttt{gpt-oss-120b} fails differently: all $24$ sessions
terminate after $1$--$2$ valid tool calls with no theory edit. \textbf{(ii)~ASP syntax.} \texttt{qwen3.5-9b} iterates but rarely emits valid clingo ($12/19$ completed samples never parse). \textbf{(iii)~CLEVRER with a reference.} DeepSeek V4 Flash reaches $100\%$ on CLEVR and $85.0\%$ on GQA $B = 0$; \texttt{qwen3.6-27b} reaches $85.0\%$ on GQA $B = 0$ and $100\%$ on its
best CLEVR sample (mean $42.2\%$, high variance). Both produce no theory for half or more of their CLEVRER $B \geq 1$ samples; \texttt{qwen3.6-27b} catalogs operators without ever writing a theory.

\section{Discussion}
\label{sec:discussion}

Three of four frontier models (RQ1) reach $100\%$ on CLEVR and $92.7\%$--$95.3\%$ on CLEVRER; GPT-5 drops to $41.8\%$ on GQA. The drop is theory coverage, not reasoning: GPT-5's GQA failures are dominated by \textsc{no\_answer} (the theory is silent on the question) rather than \textsc{semantic} (a wrong answer). 
A one-shot non-agent baseline (Appendix~\ref{app:non-agent}) sharpens this picture: GPT-5 one-shot on GQA reaches $50.2 \pm 2.3$, exceeding its agent score ($41.8 \pm 9.8$). The agent loop is net-negative for GPT-5 on GQA.

References (RQ2) move Sonnet, Opus, and DeepSeek V4 Pro within $\pm 3.4$\,pp. GPT-5 regresses on every dataset ($-9.9$, $-6.6$, $-9.4/{-3.2}$, $-19.4$\,pp on CLEVR, GQA, CLEVRER $B{=}1$ CLEVR-/GQA-ref, $B{=}2$). The mechanism is unconfirmed; one observation is that GPT-5 writes shorter theories when given a reference, so the reference may consume context budget that would otherwise go to authoring.

The capability threshold (RQ3) is not pure scale. \texttt{gpt-oss-20b} fails on tool call format; \texttt{gpt-oss-120b} fails differently (Appendix~\ref{app:failures}). Below 27B, \texttt{qwen3.5-9b} parses on only $7/19$ samples. At 27B and above, \texttt{qwen3.6-27b} and DeepSeek V4 Flash approach the frontier on CLEVR and GQA $B = 0$ but fail on CLEVRER when given a reference.

Session-level failure modes (RQ4) fire zero times for the four frontier models; the five sub-frontier models exhibit four of them (Appendix~\ref{app:failures}). Frontier failures are entirely per-val-example: three at $2.5$--$4.2\%$ total error and GPT-5 at $27.6\%$, dominated by \textsc{no\_answer} on GQA. Sub-frontier models are dominated by \textsc{no\_answer} (32--100\%); below 27B, parse failures additionally affect $33.3\%$ (qwen3.5-9b) and $4.2\%$ (gpt-oss-20b) of val examples.

\section{Conclusion}

We investigated whether an LLM agent, given an empty file and a solver in the loop, can distill a complete ASP theory from scratch. We found that three of four frontier models (Sonnet,
Opus, DeepSeek V4 Pro) author theories that score $\geq 93\%$ on every benchmark and meet or exceed the handwritten reference on GQA; below the $27$\,B-parameter boundary, models either fail to produce parseable theories or stall before authoring.
Surprisingly, the fourth frontier model (GPT-5) collapses on GQA ($41.8\%$), and a one-shot non-agent baseline (Appendix~\ref{app:non-agent}) shows
the agent loop is the active cause: GPT-5 zero-shot reaches $50.2\%$, beating its own iterative score. Next steps: plug the distilled theory into NS-CL~\cite{mao2019neuro} as the
symbolic reasoning module and train a vision module from the solver's feedback~\cite{yi2018neural}, auto-generate the question parser from each dataset's functional-program
annotations, and characterize the conditions under which the agent loop helps versus hurts.

\clearpage
\bibliography{references}

\clearpage
\appendix

\section{Models evaluated}
\label{app:models}

\begin{table}[ht]
\centering
\small
\begin{tabular}{l l r}
\toprule
Model & Provider & Size \\
\midrule
GPT-5~\cite{openai2025gpt5}                   & OpenAI    & $\sim$1\,T (est.) \\
DeepSeek V4 Pro~\cite{deepseek2025v4}         & DeepSeek  & 671\,B MoE \\
Claude Opus 4.7~\cite{anthropic2025claude}    & Anthropic & $\sim$400\,B (est.) \\
Claude Sonnet 4.6~\cite{anthropic2025claude}  & Anthropic & $\sim$150\,B (est.) \\
DeepSeek V4 Flash~\cite{deepseek2025v4}       & DeepSeek  & $\sim$30\,B (est.) \\
gpt-oss-120b                                  & OpenAI    & 120\,B \\
qwen3.6-27b                                   & Qwen      & 27\,B \\
gpt-oss-20b                                   & OpenAI    & 20\,B \\
qwen3.5-9b                                    & Qwen      & 9\,B \\
\bottomrule
\end{tabular}
\caption{Models evaluated, ordered by estimated size. GPT-5 and the Claude family have not publicly disclosed parameter counts (sizes shown are community estimates, prefixed with $\sim$); DeepSeek V4 Pro's MoE size and the open-weights models are exact.}
\label{tab:models}
\end{table}

\section{System prompt and directive}
\label{app:prompt}

The same two prompts drive every configuration. The system prompt (\texttt{prompts/system.md}), verbatim:

\begin{lstlisting}
# Task

You are an Answer Set Programming (ASP) agent. Your working directory
is the run directory.

# Goal

Produce a theory in `theory.lp` such that, for each training example,
the `clingo` solver derives the ground truth answer `ans(A)` when given
`theory.lp` combined with the example's scene facts and question facts.
Your theory will be scored on a held out validation set that is not
visible to you; iterate against the training examples until your rules
cover the training distribution.

# What is on disk

- `theory.lp` -- starts empty. You author it.
- `data/train/<idx>.json` -- training examples. Each file has
`scene_asp`, `question_asp`, `expected_answer`, `raw_question`.
- `references/` -- read only handwritten ASP theories for OTHER datasets.
May be empty. They target different schemas; do not copy them verbatim.

# Tools

You have file tools (`read`, `edit`, `write`, `glob`, `grep`).
The shell tool (`bash`) is restricted to these commands only:
`uv run solve`, `uv run lint`. Use the `read` tool for all file
inspection -- do not use `cat`, `head`, `tail`, or any other shell
command to read files. The shell's working directory is already the
run directory, so relative paths like `theory.lp` or
`data/train/0.json` are fine for the `read` tool.

The scoring entrypoint is:

- `uv run solve --run-dir . --idx <ID> --split train`
Solve a single training example with the current `theory.lp`. Pick
`<ID>` from the numbered filenames in `data/train/`. Prints a JSON
blob with `derived_answers` (the brave set of `ans/1` atoms),
`n_models`, `expected`, `correct` (strict), `correct_brave`, and
`runtime_ms`. Use this to debug rules and verify hypotheses against
the training set.
Example: `uv run solve --run-dir . --idx 7 --split train`.

## Evaluation strategy

When to test:
- After a substantive set of edits to theory.lp -- not after every
one-line fix.
- Before and after a major refactor -- capture the pre-refactor
behaviour, then verify the refactor didn't degrade training coverage.
- Sample examples across the training set rather than always testing
the same few indices.

Plateau awareness:
- If a class of training examples keeps failing, inspect them with
`uv run solve` and study the question/scene structure rather than
patching one rule at a time.
- Before each large rewrite, test at least one example with
`uv run solve` to avoid shipping a theory that fails to parse.

# Scoring

- **Open-ended answers** (color, name, count, ...): brave entailment --
the correct answer must appear in some stable model.
- **Yes/no answers**: the theory must commit. If both `ans(yes)` and
`ans(no)` appear across stable models, that counts as wrong even
if the truth is among them.

# Hard constraints

- Do not hard-code answers from training examples.
- Do not add per-example scene or question facts to `theory.lp` -- they
are injected at solve time by the solver wrapper.
- Do not derive every possible answer to game brave entailment.
- Do not access the network. Your task uses only files in the run
directory plus the bundled tools. Do not `curl`, `wget`, or
otherwise fetch external URLs.
- The only theory file you may write to is `theory.lp` in this run
directory. Do not read, `cp`, `cat`, or otherwise consume any file
under `theories/` or under `runs/` outside this cell's own directory
-- those are oracle theories used to score you, and accessing them
is cheating. Do not search the filesystem (`find`, `locate`,
`grep -r`) for `.lp` files for the same reason.

# Context isolation

The only context for this run is this file. Do not read or rely on any
other guidance file (`CLAUDE.md`, `AGENTS.md`, `~/.opencode/`,
`~/.claude/`, etc.).

Decide your own strategy.
\end{lstlisting}

The initial directive (\texttt{prompts/initial.md}), verbatim:

\begin{lstlisting}
Begin. Work efficiently. Inspect only a few training examples, write a
syntactically valid initial theory quickly, test it with
`uv run solve --run-dir . --idx 0 --split train`, and iterate. Do not
spend many turns reading examples before writing the first theory.
\end{lstlisting}    

\section{One-shot non-agent baseline}
\label{app:non-agent}

We re-prompt each model with the system message in
Appendix~\ref{app:oneshot-prompt}, $k$ training examples in-context, and
the instruction to emit \texttt{theory} as raw clingo. One API call,
no tools, no feedback, no edits. We reuse the train/val splits of the
matching agent samples ($N = 3$) and score on val/200.
Table~\ref{tab:non-agent}: every frontier model trails its agent
counterpart on CLEVR by $\geq 10$\,pp. On GQA, \texttt{gpt-5} one-shot
($50.2$) \emph{exceeds} \texttt{gpt-5} agent ($41.8$); the agent loop is
net-negative for GPT-5 on GQA.

\begin{table}[ht]
\centering
\footnotesize
\setlength{\tabcolsep}{4pt}
\begin{tabular}{@{}l l c c c c c@{}}
\toprule
Dataset & Condition & acc\_strict (\%) & \textsc{parse} & \textsc{ground} & \textsc{no\_ans} & \textsc{semantic} \\
\midrule
CLEVR & \texttt{gpt-5}, one-shot              & $ 0.0 \pm 0.0$  & 33.3 & 66.7 &  0.0 &  0.0 \\
CLEVR & \texttt{claude-sonnet-4-6}, one-shot  & $60.7 \pm 4.9$  &  0.0 &  0.0 & 15.7 & 23.7 \\
CLEVR & \texttt{claude-opus-4-7}, one-shot    & $71.5 \pm 14.3$ &  0.0 &  0.0 &  3.5 & 25.0 \\
CLEVR & \texttt{gpt-5-codex}, one-shot        & $88.7 \pm 3.7$  &  0.0 &  0.0 &  3.8 &  7.5 \\
CLEVR & \texttt{gpt-5}, agent                 & $98.7 \pm 1.9$  &  0.0 &  0.0 &  0.0 &  1.3 \\
CLEVR & \texttt{claude-sonnet-4-6}, agent     & $100\hphantom{.0} \pm 0\hphantom{.0}$ &  0.0 &  0.0 &  0.0 &  0.0 \\
CLEVR & \texttt{claude-opus-4-7}, agent       & $100\hphantom{.0} \pm 0\hphantom{.0}$ &  0.0 &  0.0 &  0.0 &  0.0 \\
\midrule
GQA   & \texttt{gpt-5}, one-shot              & $50.2 \pm 2.3$  &  0.0 &  0.0 & 21.8 & 28.0 \\
GQA   & \texttt{claude-sonnet-4-6}, one-shot  & $ 0.0 \pm 0.0$  & 66.7 & 33.3 &  0.0 &  0.0 \\
GQA   & \texttt{claude-opus-4-7}, one-shot    & $46.7 \pm 13.2$ &  0.0 &  0.0 & 46.2 &  7.2 \\
GQA   & \texttt{gpt-5}, agent                 & $41.8 \pm 9.8$  &  --- &  --- &  --- &  --- \\
GQA   & \texttt{claude-sonnet-4-6}, agent     & $98.8 \pm 0.8$  &  --- &  --- &  --- &  --- \\
GQA   & \texttt{claude-opus-4-7}, agent       & $97.5 \pm 1.2$  &  --- &  --- &  --- &  --- \\
\bottomrule
\end{tabular}
\caption{One-shot baseline vs.\ agent on CLEVR / GQA at $B = 0$,
$k = 10$, $N = 3$. Error columns are per-val-example rates; agent rows
show \texttt{acc\_strict} from Table~\ref{tab:results}.
\texttt{gpt-5-codex} was not run in the agent condition.}
\label{tab:non-agent}
\end{table}

\paragraph{Sensitivity to $k$.} Higher $k$ does not close the gap (Table~\ref{tab:k-sweep}). The
runnable samples score higher (Sonnet CLEVR best: $64 \to 73.5 \to 81$;
GPT-5 GQA best: $52.5 \to 82.5$), but more theories fail to parse or
ground. Sonnet on GQA produces no runnable theory at any $k$. The GPT-5
GQA inversion intensifies: at $k = 30$ a single sample reaches $82.5$,
double the agent's $41.8$.

\begin{table}[ht]
\centering
\footnotesize
\setlength{\tabcolsep}{4pt}
\begin{tabular}{@{}l l c c c@{}}
\toprule
Model & Dataset & $k = 10$ & $k = 30$ & $k = 50$ \\
\midrule
\texttt{claude-sonnet-4-6} & CLEVR & $60.7 \pm 4.9$ (3/3) & $44.5 \pm 38.6$ (2/3) & $27.0 \pm 46.8$ (1/3) \\
\texttt{claude-sonnet-4-6} & GQA   & $ 0.0 \pm 0.0$ (0/3) & $ 0.0 \pm 0.0$  (0/3) & $ 0.0 \pm 0.0$  (0/3) \\
\texttt{gpt-5}             & GQA   & $50.2 \pm 2.3$ (3/3) & $47.8 \pm 41.6$ (2/3) & $ 0.0 \pm 0.0$  (0/3) \\
\bottomrule
\end{tabular}
\caption{One-shot sensitivity to in-context size $k$. Mean $\pm$ std
across $N = 3$; parenthesis is the number of samples whose theory parsed
and grounded.}
\label{tab:k-sweep}
\end{table}

\section{Token usage}
\label{app:tokens}

Per agent sample, mean across configurations. \textit{in}: new input $+$
cache read $+$ cache write. \textit{out}: generated, incl.\ reasoning.

\begin{table}[ht]
\centering
\footnotesize
\setlength{\tabcolsep}{3pt}
\begin{tabular}{@{}l rr rr rr@{}}
\toprule
 & \multicolumn{2}{c}{CLEVR}
 & \multicolumn{2}{c}{GQA}
 & \multicolumn{2}{c}{CLEVRER} \\
\cmidrule(lr){2-3} \cmidrule(lr){4-5} \cmidrule(lr){6-7}
Model & in (M) & out (k) & in (M) & out (k) & in (M) & out (k) \\
\midrule
Sonnet 4.6        &  0.79 &  40.2 & 11.78 & 119.4 & 23.86 & 155.6 \\
Opus 4.7          &  0.44 &  22.8 &  6.07 &  66.0 &  9.53 &  87.6 \\
GPT-5             &  2.11 &  27.1 &  7.52 &  34.8 &  2.61 &  24.7 \\
DeepSeek V4 Pro   &  2.31 &  42.2 & 13.52 &  90.5 & 12.78 & 106.6 \\
\midrule
DeepSeek V4 Flash &  5.18 &  79.0 & 10.69 &  92.2 &  8.87 & 160.5 \\
gpt-oss-120b      &  0.01 &   0.1 &  0.01 &   0.1 &  0.01 &   0.1 \\
qwen3.6-27b       &  0.56 &  55.2 &  1.52 & 105.2 &  0.37 &  38.8 \\
gpt-oss-20b       &  0.02 &   1.8 &  0.01 &   0.1 &  0.01 &   0.3 \\
qwen3.5-9b        &  2.41 &  36.2 &  0.17 &   5.0 &  4.35 &  39.7 \\
\bottomrule
\end{tabular}
\caption{Agent token usage per sample, mean across configurations of
each (model, dataset) pair. Session totals up to the 1-hour cap or model
self-stop.}
\label{tab:tokens}
\end{table}

\begin{table}[ht]
\centering
\footnotesize
\setlength{\tabcolsep}{4pt}
\begin{tabular}{@{}l l r r@{}}
\toprule
Dataset & Model & in & out \\
\midrule
CLEVR & \texttt{gpt-5}             &   5\,732 & 18\,492 \\
CLEVR & \texttt{claude-sonnet-4-6} &   7\,187 &  2\,886 \\
CLEVR & \texttt{claude-opus-4-7}   &   7\,984 &  4\,131 \\
CLEVR & \texttt{gpt-5-codex}       &  32\,699 & 37\,249 \\
GQA   & \texttt{gpt-5}             &  41\,589 &  7\,432 \\
GQA   & \texttt{claude-sonnet-4-6} &  55\,571 &  2\,081 \\
GQA   & \texttt{claude-opus-4-7}   &  64\,357 &  1\,718 \\
\bottomrule
\end{tabular}
\caption{One-shot baseline token usage per sample at $k = 10$ (mean of
$N = 3$). Two to four orders of magnitude below the agent runs in
Table~\ref{tab:tokens}.}
\label{tab:tokens-oneshot}
\end{table}

\section{One-shot baseline prompt}
\label{app:oneshot-prompt}

System prompt, verbatim:

\begin{lstlisting}
You are an Answer Set Programming (ASP) agent.

# Goal

Produce a theory `theory.lp` such that, for each example, the `clingo`
solver derives the ground-truth answer `ans(A)` when given `theory.lp`
combined with the example's scene facts and question facts.

# Input format

Each training example has `scene_asp`, `question_asp`, `expected_answer`,
`raw_question`. The scene and question facts are injected at solve time;
do not include them in the theory.

# Scoring

- Open-ended answers (color, name, count, ...): brave entailment.
- Yes/no answers: the theory must commit to a single value.

# Hard constraints

- Do not hard-code answers from training examples.
- Do not derive every possible answer to game brave entailment.
- Output ONLY ASP source code. Your entire response must be valid clingo.
  Do not write any prose, explanation, planning, or markdown fences.
  Start the response with a `%` comment or an ASP rule.
\end{lstlisting}

The user message lists $k$ examples (\texttt{raw\_question},
\texttt{expected\_answer}, \texttt{scene\_asp}, \\\texttt{question\_asp})
and asks for a theory body covering them. Markdown-fenced output is
unwrapped: the largest \texttt{```...```} block in the response is used,
otherwise the raw response.

\section{Sample distilled rules}
\label{app:rules-sample}

The excerpt below is from the GQA distilled theory at $B = 0$ (Sonnet 4.6). It shows the model's handling of \texttt{select} when the parser passes a pronoun rather than a class name. The model renamed the handwritten reference's \texttt{state/2} predicate to \texttt{node/2} (cf.\ Figure~\ref{fig:worked-example}); the rule shape is identical. The handwritten GQA theory in the repository has equivalent but not identical handling of pronouns. Full distilled theories and handwritten baselines are in Appendices~\ref{app:theory-clevr}--\ref{app:theory-clevrer}.

\begin{lstlisting}[style=asp]
% select(S, _, Name): bind objects whose name or class matches Name
node(S, O) :- select(S, _, Name), object(O), has_attr(O, name,  Name).
node(S, O) :- select(S, _, Name), object(O), has_attr(O, class, Name).

% Pronoun resolution: "he" matches male-gendered or male-typed objects.
node(S, O) :- select(S, _, he), object(O), has_attr(O, gender, male).
node(S, O) :- select(S, _, he), object(O), has_attr(O, name,  man).
node(S, O) :- select(S, _, he), object(O), has_attr(O, name,  boy).
\end{lstlisting}

\section{Failure modes}
\label{app:failures}

We list each and note which models it fired for. Table~\ref{tab:failures} reports per-val-example rates per model.

\paragraph{Per-example modes.}
\begin{itemize}\setlength{\itemsep}{1pt}
  \item \textsc{parse}: theory does not parse in \texttt{clingo}, so no answer is derivable for this val example. \emph{Fires for smaller models}: qwen3.5-9b on $12/19$ completed samples, gpt-oss-20b on $1/24$ sample. Zero for every frontier model.
  \item \textsc{ground}: theory parses but \texttt{clingo} cannot ground it for this example (e.g., unsafe variable). \emph{Folded with} \textsc{parse} in the body table.
  \item \textsc{no\_answer}: theory parses and grounds but derives no \texttt{ans/1} atom. The model's theory is silent on this question shape.
  \item \textsc{semantic}: theory derives a value that disagrees with the ground truth.
\end{itemize}

\paragraph{Session-level modes.}
\begin{itemize}\setlength{\itemsep}{1pt}
  \item \textsc{Runaway exploration}: the model spends the 1-hour wall reading examples and reasoning without writing rules; \texttt{theory} stays empty. Zero for frontier models. Fires for qwen3.6-27b on CLEVRER $B = 1$ CLEVR-ref and $B = 2$ ($0/3$ valid theories; the model catalogs operators via its \texttt{task} subagent without authoring).
  \item \textsc{Parse-broken theory}: the final \texttt{theory} fails to parse, so every val example errors before scoring. Zero for frontier models. Fires for qwen3.5-9b (the source of the $33.3\%$ \textsc{parse} rate in Table~\ref{tab:failures}).
  \item \textsc{Tool call dialect mismatch}: model emits tool calls in a format the harness rejects. Zero for frontier models. Fires for gpt-oss-20b ($17/24$ samples emit malformed tool names the harness cannot parse). \texttt{gpt-oss-120b}'s $24/24$ failures are distinct: sessions terminate after $1$--$2$ valid tool calls with no theory edit; mechanism unclear.
  \item \textsc{Early stop}: the model emits \texttt{step\_finish} with reason \texttt{stop} after a few reasoning steps, with no edits to \texttt{theory}. Zero for frontier models. Fires for gpt-oss-20b ($6/24$ completed-but-null samples).
  \item \textsc{Soft-fail plateau}: the model edits productively for a stretch then stalls below a peer model's accuracy. Difficult to count automatically; observed qualitatively in GPT-5 on GQA (the theory plateaus around $42\%$ for many edits before the agent reports done).
\end{itemize}

\section{Handwritten theories used as references}
\label{app:hand-theories}

The handwritten ASP theories for CLEVR and GQA used as reference material in the $B \geq 1$ conditions are taken from prior work in this line: \citet{eiter2022pipeline} for CLEVR and \citet{eiter2023modular} for GQA, with the extension to contrastive explanations developed in \citet{eiter2023contrastive}. Each theory targets the corresponding dataset's question grammar from \citet{johnson2017clevr} (CLEVR) and \citet{hudson2019gqa} (GQA). Full listings are released with the code repository (\texttt{theories/handwritten/}). CLEVRER has no handwritten theory in our setup.

\section{Distilled theory: CLEVR (Opus 4.7, \texorpdfstring{$B = 0$}{B=0})}
\label{app:theory-clevr}
\label{app:dist-clevr}
The sample reproduced below reaches \texttt{acc\_strict} = 1.00 on val/200; see Table~\ref{tab:results} for the mean across $N = 3$ samples.
\begin{lstlisting}[style=asp,basicstyle=\ttfamily\scriptsize]
% CLEVR-style ASP theory
% obj(Id, X, Y, Material, Color, Shape, Size)
% Relations: left_of(A,B), right_of(A,B), behind_of(A,B), in_front_of(A,B)
% A program is a sequence of steps with positions 0,1,2,...
% Each non-binary op X(I) takes step I-1 as input.
% Binary ops:
%   or_op(I, J)    : own=I, other_branch_start=J. Inputs from I-1 (own branch end) and J-1 (other branch end)
%   and_op(I, J)   : same convention as or_op
%   equal_X(J, I)  : other_branch_start=J, own=I. Inputs from J-1 and I-1.
%   greater_than(J, I) / less_than(J, I) / equal_integer(J, I): same convention as equal_X

% ============================================================
% Step ID extraction (used to identify the final answer step)
% ============================================================
step_id(I) :- scene(I).
step_id(I) :- filter_red(I).
step_id(I) :- filter_blue(I).
step_id(I) :- filter_green(I).
step_id(I) :- filter_yellow(I).
step_id(I) :- filter_purple(I).
step_id(I) :- filter_brown(I).
step_id(I) :- filter_cyan(I).
step_id(I) :- filter_gray(I).
step_id(I) :- filter_cube(I).
step_id(I) :- filter_sphere(I).
step_id(I) :- filter_cylinder(I).
step_id(I) :- filter_small(I).
step_id(I) :- filter_large(I).
step_id(I) :- filter_rubber(I).
step_id(I) :- filter_metal(I).
step_id(I) :- unique(I).
step_id(I) :- relate_left(I).
step_id(I) :- relate_right(I).
step_id(I) :- relate_behind(I).
step_id(I) :- relate_front(I).
step_id(I) :- query_shape(I).
step_id(I) :- query_colour(I).
step_id(I) :- query_material(I).
step_id(I) :- query_size(I).
step_id(I) :- count(I).
step_id(I) :- exist(I).
step_id(I) :- same_shape(I).
step_id(I) :- same_colour(I).
step_id(I) :- same_material(I).
step_id(I) :- same_size(I).
step_id(I) :- or_op(I, _).
step_id(I) :- and_op(I, _).
step_id(I) :- equal_shape(_, I).
step_id(I) :- equal_colour(_, I).
step_id(I) :- equal_material(_, I).
step_id(I) :- equal_size(_, I).
step_id(I) :- equal_integer(_, I).
step_id(I) :- greater_than(_, I).
step_id(I) :- less_than(_, I).

% ============================================================
% Scene: load all objects
% ============================================================
step_obj(I, O) :- scene(I), obj(O,_,_,_,_,_,_).

% ============================================================
% Filters
% ============================================================
step_obj(I, O) :- filter_red(I),    step_obj(I-1, O), obj(O,_,_,_,red,_,_).
step_obj(I, O) :- filter_blue(I),   step_obj(I-1, O), obj(O,_,_,_,blue,_,_).
step_obj(I, O) :- filter_green(I),  step_obj(I-1, O), obj(O,_,_,_,green,_,_).
step_obj(I, O) :- filter_yellow(I), step_obj(I-1, O), obj(O,_,_,_,yellow,_,_).
step_obj(I, O) :- filter_purple(I), step_obj(I-1, O), obj(O,_,_,_,purple,_,_).
step_obj(I, O) :- filter_brown(I),  step_obj(I-1, O), obj(O,_,_,_,brown,_,_).
step_obj(I, O) :- filter_cyan(I),   step_obj(I-1, O), obj(O,_,_,_,cyan,_,_).
step_obj(I, O) :- filter_gray(I),   step_obj(I-1, O), obj(O,_,_,_,gray,_,_).

step_obj(I, O) :- filter_cube(I),     step_obj(I-1, O), obj(O,_,_,_,_,cube,_).
step_obj(I, O) :- filter_sphere(I),   step_obj(I-1, O), obj(O,_,_,_,_,sphere,_).
step_obj(I, O) :- filter_cylinder(I), step_obj(I-1, O), obj(O,_,_,_,_,cylinder,_).

step_obj(I, O) :- filter_small(I), step_obj(I-1, O), obj(O,_,_,_,_,_,small).
step_obj(I, O) :- filter_large(I), step_obj(I-1, O), obj(O,_,_,_,_,_,large).

step_obj(I, O) :- filter_rubber(I), step_obj(I-1, O), obj(O,_,_,rubber,_,_,_).
step_obj(I, O) :- filter_metal(I),  step_obj(I-1, O), obj(O,_,_,metal,_,_,_).

% ============================================================
% Unique: pass-through
% ============================================================
step_obj(I, O) :- unique(I), step_obj(I-1, O).

% ============================================================
% Relate
% ============================================================
step_obj(I, O) :- relate_left(I),   step_obj(I-1, O2), left_of(O, O2).
step_obj(I, O) :- relate_right(I),  step_obj(I-1, O2), right_of(O, O2).
step_obj(I, O) :- relate_behind(I), step_obj(I-1, O2), behind_of(O, O2).
step_obj(I, O) :- relate_front(I),  step_obj(I-1, O2), in_front_of(O, O2).

% ============================================================
% Same X (other objects with same attribute, excluding input)
% ============================================================
step_obj(I, O) :- same_shape(I),    step_obj(I-1, O1), obj(O1,_,_,_,_,S,_),  obj(O,_,_,_,_,S,_),  O != O1.
step_obj(I, O) :- same_colour(I),   step_obj(I-1, O1), obj(O1,_,_,_,C,_,_),  obj(O,_,_,_,C,_,_),  O != O1.
step_obj(I, O) :- same_material(I), step_obj(I-1, O1), obj(O1,_,_,M,_,_,_),  obj(O,_,_,M,_,_,_),  O != O1.
step_obj(I, O) :- same_size(I),     step_obj(I-1, O1), obj(O1,_,_,_,_,_,Sz), obj(O,_,_,_,_,_,Sz), O != O1.

% ============================================================
% Queries -> step_val
% ============================================================
step_val(I, S)  :- query_shape(I),    step_obj(I-1, O), obj(O,_,_,_,_,S,_).
step_val(I, C)  :- query_colour(I),   step_obj(I-1, O), obj(O,_,_,_,C,_,_).
step_val(I, M)  :- query_material(I), step_obj(I-1, O), obj(O,_,_,M,_,_,_).
step_val(I, Sz) :- query_size(I),     step_obj(I-1, O), obj(O,_,_,_,_,_,Sz).

% ============================================================
% Count
% ============================================================
step_val(I, N) :- count(I), N = #count{O : step_obj(I-1, O)}.

% ============================================================
% Exist
% ============================================================
has_any(I) :- exist(I), step_obj(I-1, _).
step_val(I, yes) :- exist(I), has_any(I).
step_val(I, no)  :- exist(I), not has_any(I).

% ============================================================
% or_op / and_op
% ============================================================
step_obj(I, O) :- or_op(I, J), step_obj(I-1, O).
step_obj(I, O) :- or_op(I, J), step_obj(J-1, O).
step_obj(I, O) :- and_op(I, J), step_obj(I-1, O), step_obj(J-1, O).

% ============================================================
% Equal X (yes/no)
% ============================================================
equal_shape_holds(I) :- equal_shape(J, I), step_val(J-1, V), step_val(I-1, V).
step_val(I, yes) :- equal_shape(J, I), equal_shape_holds(I).
step_val(I, no)  :- equal_shape(J, I), not equal_shape_holds(I).

equal_colour_holds(I) :- equal_colour(J, I), step_val(J-1, V), step_val(I-1, V).
step_val(I, yes) :- equal_colour(J, I), equal_colour_holds(I).
step_val(I, no)  :- equal_colour(J, I), not equal_colour_holds(I).

equal_material_holds(I) :- equal_material(J, I), step_val(J-1, V), step_val(I-1, V).
step_val(I, yes) :- equal_material(J, I), equal_material_holds(I).
step_val(I, no)  :- equal_material(J, I), not equal_material_holds(I).

equal_size_holds(I) :- equal_size(J, I), step_val(J-1, V), step_val(I-1, V).
step_val(I, yes) :- equal_size(J, I), equal_size_holds(I).
step_val(I, no)  :- equal_size(J, I), not equal_size_holds(I).

equal_integer_holds(I) :- equal_integer(J, I), step_val(J-1, V), step_val(I-1, V).
step_val(I, yes) :- equal_integer(J, I), equal_integer_holds(I).
step_val(I, no)  :- equal_integer(J, I), not equal_integer_holds(I).

% ============================================================
% greater_than / less_than (numeric)
% First arg = other_branch_start (J), second arg = own (I)
% greater_than(J, I): yes iff val(J-1) > val(I-1)
%   matches "the number of [first branch] greater than [second branch]"
% ============================================================
greater_holds(I) :- greater_than(J, I), step_val(J-1, V1), step_val(I-1, V2), V1 > V2.
step_val(I, yes) :- greater_than(J, I), greater_holds(I).
step_val(I, no)  :- greater_than(J, I), not greater_holds(I).

less_holds(I) :- less_than(J, I), step_val(J-1, V1), step_val(I-1, V2), V1 < V2.
step_val(I, yes) :- less_than(J, I), less_holds(I).
step_val(I, no)  :- less_than(J, I), not less_holds(I).

% ============================================================
% Final answer: take the step with the maximum step_id (= last non-end op).
% ============================================================
last_step(I) :- I = #max{J : step_id(J)}.

ans(V) :- last_step(I), step_val(I, V).

#show ans/1.
\end{lstlisting}

\section{Distilled theory: GQA (Sonnet 4.6, \texorpdfstring{$B = 0$}{B=0})}
\label{app:dist-gqa}
The sample reproduced below reaches \texttt{acc\_strict} = 1.00 on val/200; see Table~\ref{tab:results} for the mean across $N = 3$ samples.
\begin{lstlisting}[style=asp,basicstyle=\ttfamily\scriptsize]
% ============================================================
% VQA Theory: Evaluate question programs over scene graphs
% ============================================================

% --- Type matching: object has class T or name T ---
matches_type(O, T) :- object(O), has_attr(O, class, T).
matches_type(O, T) :- object(O), has_attr(O, name, T).

% --- Explicit supertype mappings (for inconsistent class labels) ---
% device supertype: includes common electronics not labeled as device
matches_type(O, device) :- object(O), has_attr(O, class, phone).
matches_type(O, device) :- object(O), has_attr(O, class, cell_phone).
matches_type(O, device) :- object(O), has_attr(O, class, laptop).
matches_type(O, device) :- object(O), has_attr(O, class, computer).
matches_type(O, device) :- object(O), has_attr(O, class, monitor).
matches_type(O, device) :- object(O), has_attr(O, class, screen).
matches_type(O, device) :- object(O), has_attr(O, class, tablet).
matches_type(O, device) :- object(O), has_attr(O, class, keyboard).
matches_type(O, device) :- object(O), has_attr(O, class, television).
% furniture supertype
matches_type(O, furniture) :- object(O), has_attr(O, class, coffee_table).
matches_type(O, furniture) :- object(O), has_attr(O, class, dining_table).
matches_type(O, furniture) :- object(O), has_attr(O, class, side_table).
matches_type(O, furniture) :- object(O), has_attr(O, class, end_table).
matches_type(O, furniture) :- object(O), has_attr(O, class, night_stand).
matches_type(O, furniture) :- object(O), has_attr(O, class, dresser).
matches_type(O, furniture) :- object(O), has_attr(O, class, wardrobe).
matches_type(O, furniture) :- object(O), has_attr(O, class, bookcase).
% place/water/natural environment
matches_type(O, place) :- object(O), has_attr(O, class, water).
matches_type(O, place) :- object(O), has_attr(O, class, ocean).
matches_type(O, place) :- object(O), has_attr(O, class, sea).
matches_type(O, place) :- object(O), has_attr(O, class, lake).
matches_type(O, place) :- object(O), has_attr(O, class, river).
matches_type(O, place) :- object(O), has_attr(O, class, liquid).
% toy supertype
matches_type(O, toy) :- object(O), has_attr(O, class, stuffed_animal).
matches_type(O, toy) :- object(O), has_attr(O, class, teddy_bear).
matches_type(O, toy) :- object(O), has_attr(O, class, doll).
% vehicle supertype
matches_type(O, vehicle) :- object(O), has_attr(O, class, train_car).
matches_type(O, vehicle) :- object(O), has_attr(O, class, wagon).
matches_type(O, vehicle) :- object(O), has_attr(O, class, van).
% Compound type matching (e.g., "sign" matches "traffic_sign")
matches_type(O, sign) :- object(O), has_attr(O, class, traffic_sign).
matches_type(O, sign) :- object(O), has_attr(O, class, stop_sign).
matches_type(O, sign) :- object(O), has_attr(O, name, traffic_sign).
matches_type(O, sign) :- object(O), has_attr(O, name, stop_sign).
matches_type(O, sign) :- object(O), has_attr(O, name, road_sign).
% Pronoun matching
matches_type(O, she)  :- object(O), has_attr(O, class, woman).
matches_type(O, she)  :- object(O), has_attr(O, class, girl).
matches_type(O, she)  :- object(O), has_attr(O, gender, female).
matches_type(O, he)   :- object(O), has_attr(O, class, man).
matches_type(O, he)   :- object(O), has_attr(O, class, boy).
matches_type(O, he)   :- object(O), has_attr(O, gender, male).
matches_type(O, it)   :- object(O).
matches_type(O, this) :- object(O).
matches_type(O, that) :- object(O).
matches_type(O, they) :- object(O).

% --- Plural forms for query results ---
has_attr(O, name, V2) :- object(O), has_attr(O, name, V), plural_form(V, V2).
plural_form(horse, horses). plural_form(cow, cows). plural_form(goat, goats).
plural_form(onion, onions). plural_form(topping, toppings). plural_form(sign, signs).
plural_form(shelf, shelves). plural_form(leaf, leaves). plural_form(knife, knives).
plural_form(tree, trees). plural_form(bird, birds). plural_form(dog, dogs).
plural_form(cat, cats). plural_form(car, cars). plural_form(duck, ducks).
plural_form(person, people). plural_form(child, children).
plural_form(man, men). plural_form(woman, women). plural_form(boy, boys).
plural_form(girl, girls). plural_form(bear, bears). plural_form(sheep, sheep).
plural_form(fish, fish). plural_form(deer, deer).
plural_form(flower, flowers). plural_form(banana, bananas). plural_form(apple, apples).
plural_form(pizza, pizzas). plural_form(sandwich, sandwiches).
plural_form(cup, cups). plural_form(bottle, bottles). plural_form(box, boxes).
plural_form(plate, plates). plural_form(bowl, bowls).

% --- Virtual scene object: aggregates scene-level attributes ---
has_attr(scene_obj, Attr, V) :- object(O), has_attr(O, Attr, V), is_attr(Attr).
has_attr(scene_obj, place, V) :- object(O), has_attr(O, class, place), has_attr(O, name, V).
has_attr(scene_obj, room, V)  :- object(O), has_attr(O, class, room), has_attr(O, name, V).
% Infer indoor/outdoor location from scene objects
indoor_class(wall). indoor_class(ceiling). indoor_class(floor). indoor_class(curtain).
indoor_class(carpet). indoor_class(rug). indoor_class(indoor). indoor_class(staircase).
indoor_class(stair). indoor_class(window). indoor_class(door). indoor_class(bedroom).
indoor_class(kitchen). indoor_class(bathroom). indoor_class(living_room). indoor_class(office).
has_attr(scene_obj, location, indoors) :- object(O), has_attr(O, class, C), indoor_class(C).
has_attr(scene_obj, location, indoors) :- has_attr(scene_obj, room, _).
outdoor_class(sky). outdoor_class(cloud). outdoor_class(sun). outdoor_class(outdoor).
outdoor_class(field). outdoor_class(beach). outdoor_class(park).
has_attr(scene_obj, location, outdoors) :- object(O), has_attr(O, class, C), outdoor_class(C).

% --- Helper: membership (set or single object) ---
node_member(N, O) :- node_set(N, O).
node_member(N, O) :- node_obj(N, O).

% ============================================================
% scene: root node representing the entire scene
% ============================================================
node_scene(N) :- scene(N).

% ============================================================
% select(N, M, T): select objects of type T
% ============================================================
node_scene(N) :- select(N, M, scene), node_scene(M).
node_set(N, O) :- select(N, M, T), node_scene(M), T != scene, matches_type(O, T).

% ============================================================
% unique(N, M): pick one element from set M
% Use existential semantics: all objects flow through (best for
% yes/no questions with multi-object sets)
% ============================================================
node_obj(N, O) :- unique(N, M), node_set(M, O).
node_obj(N, scene_obj) :- unique(N, M), node_scene(M).

% ============================================================
% query(N, M, Attr): get attribute value of object at node M
% ============================================================
node_val(N, V) :- query(N, M, Attr), node_obj(M, O), has_attr(O, Attr, V).

% ============================================================
% verify_attr(N, M, Attr, Val): yes if object has attr=val
% ============================================================
node_bool(N, yes) :- verify_attr(N, M, Attr, Val),
                     node_obj(M, O), has_attr(O, Attr, Val).
node_bool(N, no)  :- verify_attr(N, M, _, _),
                     node_obj(M, _), not node_bool(N, yes).

% ============================================================
% filter(N, M, Attr, Val): keep objects with attr=val
% ============================================================
node_set(N, O) :- filter(N, M, Attr, Val),
                  node_set(M, O), has_attr(O, Attr, Val).

% ============================================================
% filter_any(N, M, Val): keep objects with has_attr(O, any, Val)
% ============================================================
node_set(N, O) :- filter_any(N, M, Val),
                  node_set(M, O), has_attr(O, any, Val).

% ============================================================
% negate(N, M1, M2): objects in M2 but not M1
% ============================================================
node_set(N, O) :- negate(N, M1, M2),
                  node_set(M2, O), not node_set(M1, O).

% ============================================================
% exist(N, M): yes if M is non-empty
% ============================================================
node_bool(N, yes) :- exist(N, M), node_set(M, _).
node_bool(N, no)  :- exist(N, M), not node_bool(N, yes).

% ============================================================
% and / or
% ============================================================
node_bool(N, yes) :- and(N, A, B), node_bool(A, yes), node_bool(B, yes).
node_bool(N, no)  :- and(N, A, B), not node_bool(N, yes).

node_bool(N, yes) :- or(N, A, B), node_bool(A, yes).
node_bool(N, yes) :- or(N, A, B), node_bool(B, yes).
node_bool(N, no)  :- or(N, A, B), not node_bool(N, yes).

% ============================================================
% relate(N, M, Type, Rel, Role): find objects of Type via Rel
%   subject: find T where has_rel(T, Rel, O)
%   object:  find T where has_rel(O, Rel, T)
% ============================================================
node_set(N, T) :- relate(N, M, Type, Rel, subject),
                  node_member(M, O),
                  has_rel(T, Rel, O), matches_type(T, Type).
node_set(N, T) :- relate(N, M, Type, Rel, object),
                  node_member(M, O),
                  has_rel(O, Rel, T), matches_type(T, Type).

% ============================================================
% relate_any(N, M, Rel, Role): find related objects
%   4th arg is ROLE (subject/object), not a type
%   object:  M is subject, find T where has_rel(O, Rel, T)
%   subject: M is object,  find T where has_rel(T, Rel, O)
% ============================================================
node_set(N, T) :- relate_any(N, M, Rel, object),
                  node_member(M, O), has_rel(O, Rel, T), object(T).
node_set(N, T) :- relate_any(N, M, Rel, subject),
                  node_member(M, O), has_rel(T, Rel, O), object(T).

% ============================================================
% choose_attr(N, M, Attr, V1, V2)
% Also derive plural forms for completeness
% ============================================================
node_val(N, V1) :- choose_attr(N, M, Attr, V1, _),
                   node_obj(M, O), has_attr(O, Attr, V1).
node_val(N, V2) :- choose_attr(N, M, Attr, _, V2),
                   node_obj(M, O), has_attr(O, Attr, V2).
% Also derive plurals of chosen values
node_val(N, Vp) :- choose_attr(N, M, Attr, V1, _),
                   node_obj(M, O), has_attr(O, Attr, V1), plural_form(V1, Vp).
node_val(N, Vp) :- choose_attr(N, M, Attr, _, V2),
                   node_obj(M, O), has_attr(O, Attr, V2), plural_form(V2, Vp).

% ============================================================
% choose_rel(N, M, TgtType, R1, R2, Role)
% ============================================================
node_val(N, Label) :- choose_rel(N, M, TgtType, R1, R2, object),
                      node_obj(M, O), matches_type(T, TgtType),
                      has_rel(O, R1, T), rel_label(R1, Label).
node_val(N, Label) :- choose_rel(N, M, TgtType, R1, R2, object),
                      node_obj(M, O), matches_type(T, TgtType),
                      has_rel(O, R2, T), rel_label(R2, Label).
node_val(N, Label) :- choose_rel(N, M, TgtType, R1, R2, subject),
                      node_obj(M, O), matches_type(T, TgtType),
                      has_rel(T, R1, O), rel_label(R1, Label).
node_val(N, Label) :- choose_rel(N, M, TgtType, R1, R2, subject),
                      node_obj(M, O), matches_type(T, TgtType),
                      has_rel(T, R2, O), rel_label(R2, Label).

rel_label(to_the_left_of, left).
rel_label(to_the_right_of, right).
rel_label(R, R) :- choose_rel(_, _, _, R, _, _), R != to_the_left_of, R != to_the_right_of.
rel_label(R, R) :- choose_rel(_, _, _, _, R, _), R != to_the_left_of, R != to_the_right_of.

% ============================================================
% verify_rel(N, M, TgtType, Rel, Role)
% ============================================================
node_bool(N, yes) :- verify_rel(N, M, TgtType, Rel, object),
                     node_obj(M, O), matches_type(T, TgtType),
                     has_rel(O, Rel, T).
node_bool(N, no)  :- verify_rel(N, M, _, _, object),
                     node_obj(M, _), not node_bool(N, yes).

node_bool(N, yes) :- verify_rel(N, M, TgtType, Rel, subject),
                     node_obj(M, O), matches_type(T, TgtType),
                     has_rel(T, Rel, O).
node_bool(N, no)  :- verify_rel(N, M, _, _, subject),
                     node_obj(M, _), not node_bool(N, yes).

% ============================================================
% common(N, M1, M2): find attribute type shared by both objects
% ============================================================
node_val(N, Attr) :- common(N, M1, M2),
                     node_obj(M1, O1), node_obj(M2, O2),
                     has_attr(O1, Attr, _), has_attr(O2, Attr, _),
                     is_attr(Attr).

% ============================================================
% two_different(N, M1, M2, Attr): yes if attr values differ
% ============================================================
node_bool(N, yes) :- two_different(N, M1, M2, Attr),
                     node_obj(M1, O1), node_obj(M2, O2),
                     has_attr(O1, Attr, V1), has_attr(O2, Attr, V2), V1 != V2.
node_bool(N, no)  :- two_different(N, M1, M2, _),
                     node_obj(M1, _), node_obj(M2, _),
                     not node_bool(N, yes).

% ============================================================
% two_same(N, M1, M2, Attr): yes if attr values are the same
% ============================================================
node_bool(N, yes) :- two_same(N, M1, M2, Attr),
                     node_obj(M1, O1), node_obj(M2, O2),
                     has_attr(O1, Attr, V), has_attr(O2, Attr, V).
node_bool(N, no)  :- two_same(N, M1, M2, _),
                     node_obj(M1, _), node_obj(M2, _),
                     not node_bool(N, yes).

% ============================================================
% compare(N, M1, M2, Attr, true): return M1; false: return M2
% (simplified: true=return first object, false=return second)
% ============================================================
node_obj(N, O) :- compare(N, M1, _, _, true),  node_obj(M1, O).
node_obj(N, O) :- compare(N, _, M2, _, false), node_obj(M2, O).

% ============================================================
% all_same(N, M, Attr): yes if all objects in M have same Attr
% For class/name, compare by name (most specific type)
% ============================================================
all_same_val(N, V) :- all_same(N, M, class), node_set(M, O), has_attr(O, name, V).
all_same_val(N, V) :- all_same(N, M, name),  node_set(M, O), has_attr(O, name, V).
all_same_val(N, V) :- all_same(N, M, Attr),  node_set(M, O), has_attr(O, Attr, V),
                      Attr != class, Attr != name.
all_same_conflict(N) :- all_same_val(N, V1), all_same_val(N, V2), V1 != V2.
node_bool(N, yes) :- all_same(N, M, _), node_set(M, _), not all_same_conflict(N).
node_bool(N, no)  :- all_same(N, M, _), all_same_conflict(N).

% ============================================================
% Answer extraction
% ============================================================
ans(V) :- end(N), node_val(N, V).
ans(V) :- end(N), node_bool(N, V).
\end{lstlisting}

\section{Distilled theory: CLEVRER (Opus 4.7, \texorpdfstring{$B = 0$}{B=0})}
\label{app:theory-clevrer}
\label{app:dist-clevrer}
The sample reproduced below reaches \texttt{acc\_strict} = 0.97 on val/200; see Table~\ref{tab:results} for the mean across $N = 3$ samples.
\begin{lstlisting}[style=asp,basicstyle=\ttfamily\scriptsize]
% CLEVRER-style theory: interprets question_asp DSL over scene_asp facts.
% Produces ans/1 atoms.

% ---------- Scene helpers ----------
collision_evt(E) :- ev_type(E,collision).
in_evt(E)        :- ev_type(E,in).
out_evt(E)       :- ev_type(E,out).
start_evt(E)     :- ev_type(E,start).
end_evt(E)       :- ev_type(E,end).

observable_evt(E) :- collision_evt(E), not unseen(E).
observable_evt(E) :- in_evt(E), not unseen(E).
observable_evt(E) :- out_evt(E), not unseen(E).

% Direct causal-style link: shares an object and earlier frame.
direct_link(A,B) :- ev_obj(A,O), ev_obj(B,O), ev_frame(A,FA), ev_frame(B,FB), FA < FB, A != B.
ancestor_of(A,B) :- direct_link(A,B).
ancestor_of(A,C) :- direct_link(A,B), ancestor_of(B,C).

% Object descendant of (involves O at some ancestor or itself)
desc_of_obj(E,O) :- event(E), ev_obj(E,O).
desc_of_obj(E,O) :- event(E), ev_obj(A,O), ancestor_of(A,E).

% ---------- Set kind tagging ----------
set_kind_obj(I) :- scene_objects(I).
set_kind_obj(I) :- filter_color(I,_,_).
set_kind_obj(I) :- filter_material(I,_,_).
set_kind_obj(I) :- filter_shape(I,_,_).
set_kind_obj(I) :- filter_moving(I,_,_).
set_kind_obj(I) :- filter_stationary(I,_,_).
set_kind_obj(I) :- unique_obj(I,_).
set_kind_obj(I) :- get_col_partner(I,_,_).
set_kind_obj(I) :- get_object(I,_).

set_kind_evt(I) :- scene_events(I).
set_kind_evt(I) :- scene_all_events(I).
set_kind_evt(I) :- scene_unseen_events(I).
set_kind_evt(I) :- filter_collision(I,_,_).
set_kind_evt(I) :- filter_in(I,_,_).
set_kind_evt(I) :- filter_out(I,_,_).
set_kind_evt(I) :- filter_ancestor(I,_,_).
set_kind_evt(I) :- filter_after(I,_,_).
set_kind_evt(I) :- filter_before(I,_,_).
set_kind_evt(I) :- filter_order(I,_,_).
set_kind_evt(I) :- get_counterfact(I,_,_).
set_kind_evt(I) :- unique_evt(I,_).

% ---------- Object sets ----------
obj_set(I,O) :- scene_objects(I), object(O).

obj_set(I,O) :- filter_color(I,IN,C),    obj_set(IN,O), has_color(O,C).
obj_set(I,O) :- filter_material(I,IN,M), obj_set(IN,O), has_material(O,M).
obj_set(I,O) :- filter_shape(I,IN,S),    obj_set(IN,O), has_shape(O,S).

obj_set(I,O) :- unique_obj(I,IN), obj_set(IN,O).
obj_uniq(I,O) :- unique_obj(I,IN), obj_set(IN,O).

% get_col_partner: partner of given object in given collision
obj_set(I,P)  :- get_col_partner(I,EVT,OBJ),
                 evt_uniq(EVT,E), obj_uniq(OBJ,O),
                 ev_obj(E,P), P != O.
obj_uniq(I,P) :- get_col_partner(I,EVT,OBJ),
                 evt_uniq(EVT,E), obj_uniq(OBJ,O),
                 ev_obj(E,P), P != O.

% get_object: object that participates in an in/out event
obj_set(I,O)  :- get_object(I,EVT), evt_uniq(EVT,E), ev_obj(E,O).
obj_uniq(I,O) :- get_object(I,EVT), evt_uniq(EVT,E), ev_obj(E,O).

% filter_moving / filter_stationary
obj_set(I,O) :- filter_moving(I,IN,frame(F_IDX)),
                obj_set(IN,O), frame_val(F_IDX,F), moving(O,F).

obj_set(I,O) :- filter_stationary(I,IN,frame(F_IDX)),
                obj_set(IN,O), frame_val(F_IDX,F),
                visible(O,F), not moving(O,F).

obj_set(I,O) :- filter_stationary(I,IN,null),
                obj_set(IN,O), never_moves(O).

obj_set(I,O) :- filter_moving(I,IN,null),
                obj_set(IN,O), moving_ever(O).

% ---------- Event sets ----------
evt_set(I,E) :- scene_events(I),     observable_evt(E).
evt_set(I,E) :- scene_all_events(I), event(E).
% scene_unseen_events: future predicted collisions
evt_set(I,E) :- scene_unseen_events(I), unseen(E).

evt_set(I,E) :- filter_collision(I,EIN,OIN),
                evt_set(EIN,E), collision_evt(E),
                obj_set(OIN,O), ev_obj(E,O).

evt_set(I,E) :- filter_in(I,EIN,OIN),
                evt_set(EIN,E), in_evt(E),
                obj_set(OIN,O), ev_obj(E,O).

evt_set(I,E) :- filter_out(I,EIN,OIN),
                evt_set(EIN,E), out_evt(E),
                obj_set(OIN,O), ev_obj(E,O).

evt_set(I,E)  :- unique_evt(I,IN), evt_set(IN,E).
evt_uniq(I,E) :- unique_evt(I,IN), evt_set(IN,E).

% filter_ancestor: events in EIN that are ancestors of ETARGET
evt_set(I,E) :- filter_ancestor(I,EIN,ET),
                evt_set(EIN,E), evt_uniq(ET,T),
                ancestor_of(E,T).

% filter_after / filter_before: by frame
evt_set(I,E) :- filter_after(I,EIN,ER),
                evt_set(EIN,E), evt_uniq(ER,R),
                ev_frame(E,FE), ev_frame(R,FR), FE > FR.

evt_set(I,E) :- filter_before(I,EIN,ER),
                evt_set(EIN,E), evt_uniq(ER,R),
                ev_frame(E,FE), ev_frame(R,FR), FE < FR.

% get_counterfact: events that would still happen if OBJ removed
% Approx: events not descendants (through ancestor chain) of OBJ
evt_set(I,E) :- get_counterfact(I,EALL,OREM),
                evt_set(EALL,E), obj_uniq(OREM,O),
                not desc_of_obj(E,O).

% filter_order: first / second / last by frame
ev_in(I,E,F) :- filter_order(I,_,_), evt_set(I_PREV,E), ev_frame(E,F),
                filter_order(I,I_PREV,_).
% Better: define directly
order_in(I,E,F) :- filter_order(I,EIN,_), evt_set(EIN,E), ev_frame(E,F).

% first
evt_set(I,E)  :- filter_order(I,_,first), order_in(I,E,F),
                 F = #min{ FF : order_in(I,_,FF) }.
evt_uniq(I,E) :- filter_order(I,_,first), order_in(I,E,F),
                 F = #min{ FF : order_in(I,_,FF) }.

% last
evt_set(I,E)  :- filter_order(I,_,last), order_in(I,E,F),
                 F = #max{ FF : order_in(I,_,FF) }.
evt_uniq(I,E) :- filter_order(I,_,last), order_in(I,E,F),
                 F = #max{ FF : order_in(I,_,FF) }.

% second: rank by frame
order_rank(I,E,R) :- filter_order(I,_,second), order_in(I,E,F),
                     R = #count{ FF, EE : order_in(I,EE,FF), FF < F }.
evt_set(I,E)  :- filter_order(I,_,second), order_rank(I,E,1).
evt_uniq(I,E) :- filter_order(I,_,second), order_rank(I,E,1).

% ---------- Frames ----------
frame_val(I,F) :- get_frame_lit(I,start), ev_type(E,start), ev_frame(E,F).
frame_val(I,F) :- get_frame_lit(I,end),   ev_type(E,end),   ev_frame(E,F).
frame_val(I,F) :- get_frame_evt(I,EVT), evt_uniq(EVT,E), ev_frame(E,F).

% ---------- Queries ----------
str_val(I,C) :- query_color(I,IN),    obj_uniq(IN,O), has_color(O,C).
str_val(I,M) :- query_material(I,IN), obj_uniq(IN,O), has_material(O,M).
str_val(I,S) :- query_shape(I,IN),    obj_uniq(IN,O), has_shape(O,S).

% count_set: obj or evt
int_val(I,N) :- count_set(I,IN), set_kind_obj(IN), N = #count{ O : obj_set(IN,O) }.
int_val(I,N) :- count_set(I,IN), set_kind_evt(IN), N = #count{ E : evt_set(IN,E) }.

% exist_set
has_obj(I)   :- obj_set(I,_).
has_evt(I)   :- evt_set(I,_).
bool_val(I,yes) :- exist_set(I,IN), set_kind_obj(IN), has_obj(IN).
bool_val(I,no)  :- exist_set(I,IN), set_kind_obj(IN), not has_obj(IN).
bool_val(I,yes) :- exist_set(I,IN), set_kind_evt(IN), has_evt(IN).
bool_val(I,no)  :- exist_set(I,IN), set_kind_evt(IN), not has_evt(IN).

% belong_to(I, X, Y): X (obj_uniq or evt_uniq) in Y (some set)
% Standard: obj-in-obj-set, evt-in-evt-set.
% Cross-kind (obj X with evt_set Y): obj "participates in" some event of Y
%   — used for "responsible for"/"ancestor"/"counterfact" questions.
belong_pos(I) :- belong_to(I,X,Y), obj_uniq(X,O), obj_set(Y,O).
belong_pos(I) :- belong_to(I,X,Y), evt_uniq(X,E), evt_set(Y,E).
belong_pos(I) :- belong_to(I,X,Y), obj_uniq(X,O), evt_set(Y,E), ev_obj(E,O).
belong_pos(I) :- belong_to(I,X,Y), evt_uniq(X,E), obj_set(Y,O), ev_obj(E,O).
bool_val(I,yes) :- belong_pos(I).
bool_val(I,no)  :- belong_to(I,_,_), not belong_pos(I).

% negate_bool
bool_val(I,no)  :- negate_bool(I,IN), bool_val(IN,yes).
bool_val(I,yes) :- negate_bool(I,IN), bool_val(IN,no).

% ---------- Final answer ----------
ans(X) :- end(I), str_val(I,X).
ans(X) :- end(I), int_val(I,X).
ans(X) :- end(I), bool_val(I,X).
\end{lstlisting}

\end{document}